\pdfoutput=1

\documentclass[11pt]{article}

\usepackage[final]{acl}
\usepackage[utf8]{inputenc}
\usepackage[T1]{fontenc}
\usepackage{times}
\usepackage{latexsym}
\usepackage[T1]{fontenc}
\usepackage[utf8]{inputenc}
\usepackage{subcaption}
\usepackage{microtype}
\usepackage{inconsolata}
\usepackage{graphicx}
\usepackage{amsmath}
\usepackage{amssymb}
\usepackage{algorithm}
\usepackage{algorithmic}
\usepackage{tcolorbox}
\tcbuselibrary{listings,skins,breakable}
\usepackage{float}
\usepackage{xcolor}
\usepackage{xspace}  
\usepackage{enumitem} 
\newcommand{\framework}{\textsc{PhantomCircuit}\xspace}
\usepackage{tabularray} % 关键宏包
\usepackage{varwidth}
\usepackage{pifont}

\usepackage{graphicx}   % 提供 \resizebox
\usepackage{booktabs}   % 提供 \toprule,\midrule,\bottomrule,\cmidrule
\usepackage{multirow}   % 提供 \multirow
\usepackage{array}
\usepackage{tabularx}

\definecolor{green2}{RGB}{90, 130, 60}
\definecolor{lightgray}{RGB}{211, 211, 211}
\definecolor{red2}{RGB}{178, 34, 34}
\definecolor{lightblue}{RGB}{204, 229, 255} % 浅蓝色
\definecolor{lightgreen}{RGB}{204, 255, 204} % 浅绿色

\usepackage{hyperref}
\usepackage{bm}
\title{Pierce the Mists, Greet the Sky: \\Decipher Knowledge Overshadowing via Knowledge Circuit Analysis}

\author{Haoming Huang\textsuperscript{\rm 1,3,}\footnotemark[1], Yibo Yan\textsuperscript{\rm 1,\rm 2,}\footnotemark[1], Jiahao Huo\textsuperscript{\rm 1}, \\ 
\textbf{Xin Zou}\textsuperscript{\rm 1}, 
\textbf{Xinfeng Li}\textsuperscript{\rm 4},
\textbf{Kun Wang}\textsuperscript{\rm 4},
\textbf{Xuming Hu}\textsuperscript{\rm 1,\rm 2,}\footnotemark[2]\\
\\
\textsuperscript{\rm 1}The Hong Kong University of Science and Technology (Guangzhou)\\
\textsuperscript{\rm 2}The Hong Kong University of Science and Technology\\
\textsuperscript{\rm 3}Shanghai Jiao Tong University \;
\textsuperscript{\rm 4}Nanyang Technological University\\
\texttt{\{hmhuang04, yanyibo70\}@gmail.com}, 
\texttt{\{xuminghu\}@hkust-gz.edu.cn}
}

\begin{document}
\maketitle

\renewcommand{\thefootnote}{\fnsymbol{footnote}}
\footnotetext[1]{Co-first authors with equal contribution.}
\footnotetext[2]{Corresponding author.}
\renewcommand{\thefootnote}{\arabic{footnote}}

\begin{abstract}

Large Language Models (LLMs), despite their remarkable capabilities, are hampered by hallucinations. A particularly challenging variant, \textbf{knowledge overshadowing}, occurs when one piece of activated knowledge inadvertently masks another relevant piece, leading to erroneous outputs even with high-quality training data. Current understanding of overshadowing is \textit{largely confined to inference-time observations, lacking deep insights into its origins and internal mechanisms during model training}. Therefore, we introduce \textbf{\framework, a novel framework designed to comprehensively analyze and detect knowledge overshadowing}. By innovatively employing knowledge circuit analysis, \framework dissects the function of key components in the circuit and how the attention pattern dynamics contribute to the overshadowing phenomenon and its evolution throughout the training process. Extensive experiments demonstrate \framework’s effectiveness in identifying such instances, offering novel insights into this elusive hallucination and providing the research community with a new methodological lens for its potential mitigation. Our code can be found in \href{https://github.com/halfmorepiece/PhantomCircuit} {https://github.com/halfmorepiece/PhantomCircuit}.

\end{abstract}

\section{Introduction}

Large Language Models (LLMs) have witnessed explosive growth in recent years, demonstrating remarkable capabilities across a multitude of domains, including natural language understanding, generation, reasoning, and even cross-modal tasks \cite{chang2024survey,zhao2024explainability,yan2025position,yan2024survey,xun2025rtv}. Their proficiency has catalyzed transformative advancements in various applications. However, a persistent challenge that tempers their widespread adoption and reliability is the phenomenon of hallucination. Broadly, hallucinations refer to instances where models generate content that is factually incorrect, nonsensical, or unfaithful to the provided source context, despite appearing coherent and fluent \cite{rawte2023survey,chakraborty2025hallucination,wang2025comprehensive}.

\begin{figure}[t]
    \centering
    \includegraphics[width=\linewidth]{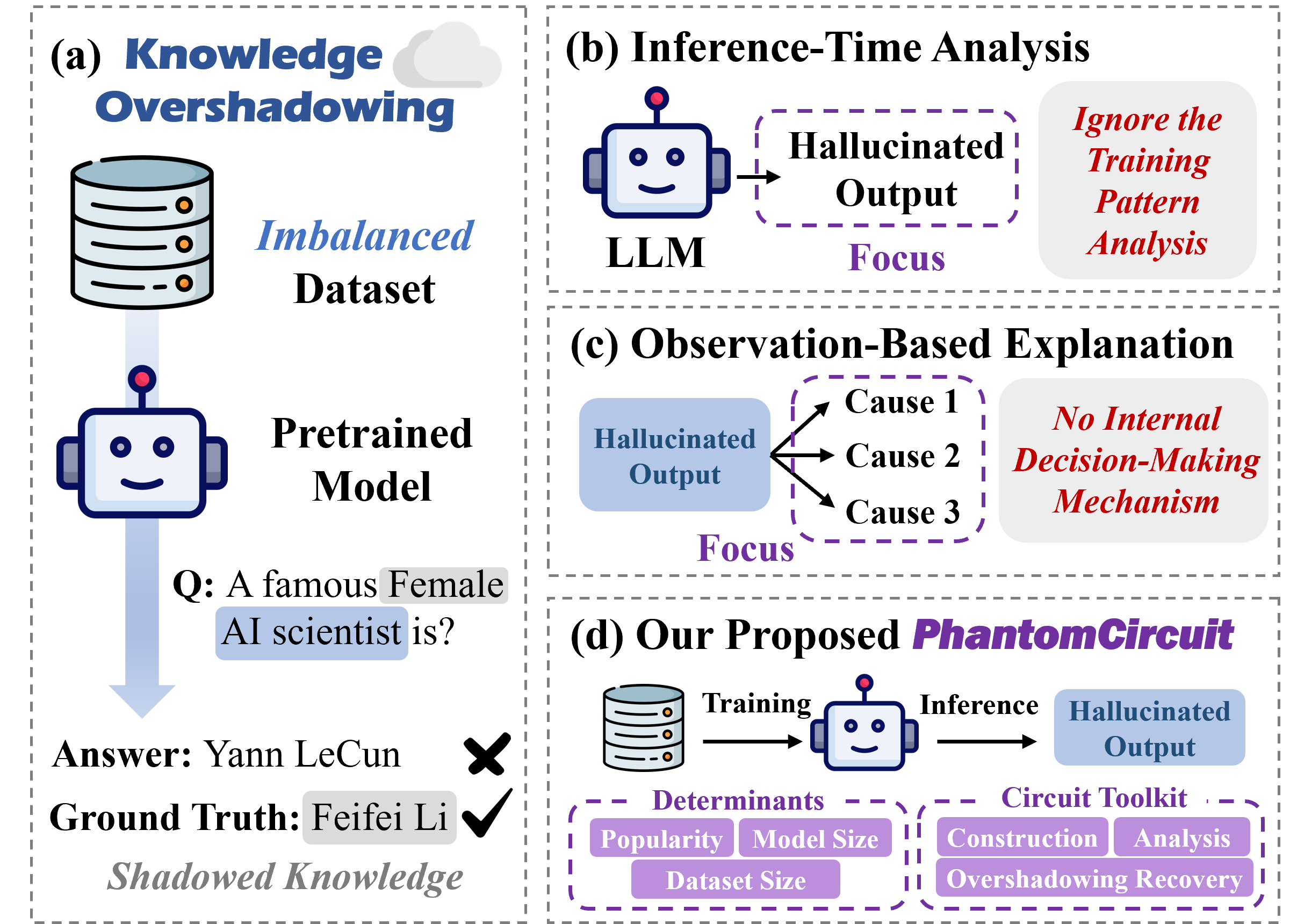}
    \caption{Illustrative comparison of previous research with inference-time analysis (b) and observation-based explanation (c) vs our proposed \framework (d) on knowledge overshadowing (a).}
    \label{fig:intro_fig}
    \vspace{-0.5cm}
\end{figure}

While substantial research has been dedicated to understanding the causes and detection of general hallucinations, a specific variant known as “\textbf{knowledge overshadowing}” warrants deeper investigation \cite{zhang2024knowledge,zhang2025law}. This phenomenon is particularly perplexing because it can manifest even when models are trained on high-quality, meticulously curated pre-training corpora. Current understanding, primarily derived from inference-time observations, characterizes overshadowing as a scenario where, for a given query, one piece of activated knowledge inadvertently “overshadows” another relevant knowledge. This interference ultimately biases the model’s reasoning process, leading to a hallucinatory output, as illustrated in Figure \ref{fig:intro_fig} (a).

Nevertheless, existing explorations into knowledge overshadowing suffer from notable limitations. \ding{182} They \textit{predominantly focus on inference-time analysis}, as shown in Figure \ref{fig:intro_fig} (b). While valuable for identifying the occurrence of overshadowing, such observations offer a surface-level understanding and often fall short of elucidating how these detrimental patterns are learned during the training phase. \ding{183} The explanations for overshadowing are often \textit{speculatively inferred from these observational outcomes} rather than being rigorously investigated through dedicated interpretability tools that can probe the model’s internal decision-making mechanisms, as shown in Figure \ref{fig:intro_fig} (c). Consequently, a more comprehensive analytical framework is imperative to dissect this phenomenon from its origins to its manifestation.

To bridge this gap, we introduce \textbf{\framework, a novel framework designed to comprehensively analyze and detect the knowledge overshadowing phenomenon}. Specifically, \framework facilitates an in-depth examination of the evolution of overshadowing hallucinations throughout the training process, correlating their emergence and prevalence with core factors such as knowledge popularity, model size, and dataset size.
Then, by leveraging knowledge circuit analysis as a key interpretability technique, we aim to trace the flow of information and the formation of knowledge representations within attention heads, thereby uncovering the internal mechanisms giving rise to overshadowing.
Furthermore, we propose to optimize the number of edges within these circuits, thus alleviating the knowledge overshadowing. As illustrated in Figure \ref{fig:intro_fig} (d), our overall work aims to provide a clearer, mechanistic understanding and potential strategies for this elusive type of hallucination.

Our contributions can be summarized as follows:

\begin{itemize}[leftmargin=*]
    \item We introduce \framework, the first comprehensive framework designed to systematically analyze and detect knowledge overshadowing, delving into its mechanistic nature and evolution throughout model training.
    \item We pioneer the use of knowledge circuit analysis to dissect the internal workings of attention heads, specifically elucidating how competing knowledge pathways contribute to the overshadowing phenomenon.
    \item We conduct extensive experiments to demonstrate \framework’s efficacy in detecting knowledge overshadowing, offering novel insights and a new methodological lens for the research community.
 
\end{itemize}

\section{Related Work}

\subsection{Hallucination Detection}

Factuality hallucination detection, which aims to evaluate whether the output of LLMs aligns with real-world facts, typically involves either external fact-checking or internal uncertainty analysis \cite{huang2025survey,dang2024exploring,zheng2024reefknot,zou2024look,zhou2024mitigating,zhu2024fastmem}. For instance, FACTSCORE~\cite{min2023factscore} decomposes a generation into atomic facts and calculates the proportion that are supported by reliable knowledge sources. FACTOOL~\cite{chern2023factool}, on the other hand, integrates multiple tools such as Google Search and Google Scholar to gather external evidence and assess the factuality of generated content. In contrast, methods like Chain-of-Verification~\cite{dhuliawala2023chain}, probability-based assessments~\cite{kadavath2022language, zhang2024self}, and uncertainty estimation approaches~\cite{varshney2023stitch, yao2023llm, luo2023zero} rely on LLMs' internal parametric knowledge or uncertainty signals to predict potential hallucinations. Among these efforts, knowledge overshadowing~\cite{zhang2025law} offers a novel perspective by modeling hallucination behavior from the perspective of knowledge representation, providing an efficient strategy for proactive prevention.

\subsection{Knowledge Circuit Analysis}

In the context of mechanistic interpretability~\cite{rai2024practical,huo2024mmneuron,huang2024miner}, computations in Transformer-based language models are viewed as a connected directed acyclic graph encompassing components such as MLPs and attention layers~\cite{syed2023attribution,conmy2023towards,huang2024inversionview}. A \textbf{circuit} refers to a sparse computational subgraph that significantly influences the model’s behavior on a specific task~\cite{olah2020zoom,elhage2021mathematical,wang2022interpretability}. Building on this, \citet{yao2024knowledge} introduce the concept of \textbf{knowledge circuits}, hypothesizing that cooperation among components reveals implicit knowledge in LLMs. Further, \citet{ou2025llms, zucchet2025languagemodelslearnfacts, hakimi2025timecoursemechinterpanalyzing} explore how such circuits evolve during continual pre-training, providing insights into knowledge acquisition. To enable effective knowledge editing, CaKE~\cite{yao2024knowledge} proposes a Circuit-aware Knowledge Editing method that guides models to activate modified knowledge and form new reasoning pathways. In this paper, we analyze the phenomenon of knowledge overshadowing through the lens of knowledge circuits, contributing new perspectives to LLM hallucination detection.

See more related work in Appendix \ref{app:llm}.

\begin{figure*}[t]
    \centering
    \includegraphics[width=0.9\textwidth]{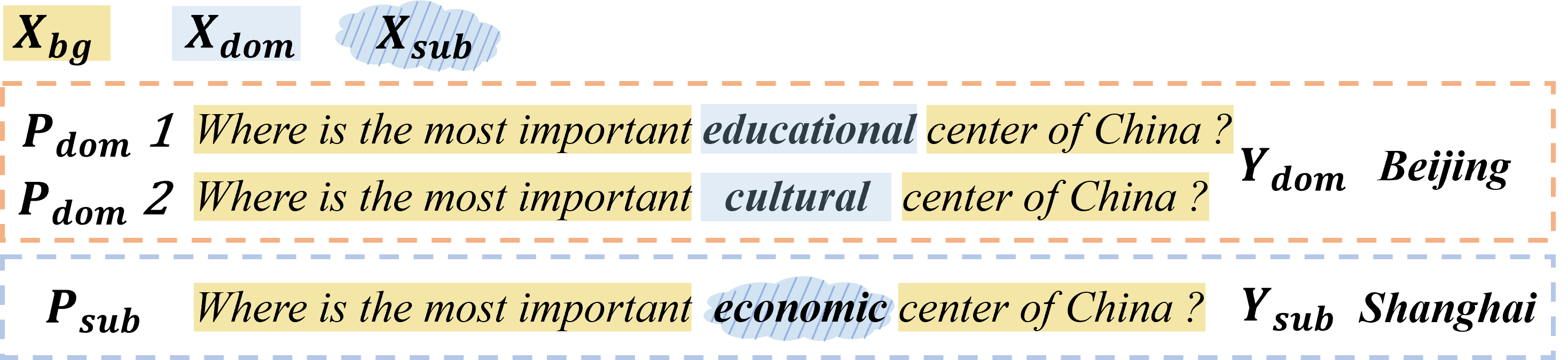}
    \caption{A group of imbalanced natural language data of knowledge overshadowing. The $X_{bg}$, $X_{dom}$ and $X_{sub}$ are background, dominant and subordinate knowledge respectively. The $P_{dom}$ and $P_{sub}$ are prompt with dominant and subordinate knowledge respectively,  their corresponding answers are $Y_{dom}$ and $Y_{sub}$.}

    \label{fig:sec3case}
     \vspace{-0.5cm}
\end{figure*}

\section{Methodology}

This section first defines the knowledge overshadowing phenomenon and its quantitative evaluation. Subsequently, we detail the \framework framework, which encompasses methods for analyzing its training dynamics and for constructing and analyzing knowledge circuits to understand its internal mechanisms\footnote{Our code will be available upon acceptance.}.

\subsection{Knowledge Overshadowing}

Knowledge overshadowing refers to a specific type of hallucination where \textit{less prevalent, subordinate knowledge} is suppressed by \textit{high-frequency, dominant knowledge} when both are associated with \textit{common background knowledge} \cite{zhang2025law}.

As shown in Figure \ref{fig:sec3case}, let $X_{dom}$ denote dominant knowledge entities and $X_{sub}$ denote subordinate knowledge entities, both potentially co-occurring with background knowledge $X_{bg}$. The core idea is that a strong learned pattern $X_{dom} \leftrightarrow X_{bg}$ can "over-generalize" where the model primarily associates $X_{dom}$ with $X_{bg}$. Consequently, when the model encounters $X_{sub}$ with $X_{bg}$, denoted as $X_{sub} \leftrightarrow X_{bg}$, it may erroneously favor outputs related to $X_{dom}$ due to the stronger $X_{dom} \leftrightarrow X_{bg}$ pattern.

\subsubsection{Knowledge Overshadowing Occurrence} 

When the input prompt is a knowledge pair composed of background knowledge and dominant knowledge, denoted as $P_{dom} = (X_{bg}, X_{dom})$, and the model correctly generates the answer corresponding to the dominant knowledge, denoted as $Y_{dom}$, we consider this outcome, represented by the pair $(P_{dom}, Y_{dom})$, as a successful recall of dominant knowledge.

When the input prompt is $P_{sub}$, but the model wrongly generates $Y_{dom}$, knowledge overshadowing occurs for this query-response instance, resulting in the $(P_{sub}, Y_{dom})$.
In the case shown in Figure \ref{fig:sec3case}, if overshadowing occurs, the model will incorrectly answer "Beijing" to the input prompt "Where is the most important economic center of China?".

\subsubsection{Quantitative Indicators}

Let $N_{dom}$ be the number of instances of the dominant knowledge prompt $P_{dom}$ and $N_{sub}$ be the number of instances of the subordinate knowledge prompt $P_{sub}$ in a group of data, respectively. In the case shown in Figure \ref{fig:sec3case}, $N_{dom}$ and $N_{sub}$ are 2 and 1, respectively. The entire dataset might contain multiple different groups of imbalanced data.

During autoregressive generation tasks performed by the model, let $M_{sub}$ be the number of times when overshadowing instances $(P_{sub}, Y_{dom})$ occur, and $M_{dom}$ be the number of times $(P_{dom}, Y_{dom})$ occurs. Then, we can define the absolute extent of the knowledge overshadowing effect, the Absolute Overshadowing rate ($\mathcal{AO}$) and calculate it using $M_{sub}$ and $N_{sub}$,

\begin{equation}
    \mathcal{AO} = p(Y_{dom}|P_{sub}) = \frac{M_{sub}}{N_{sub}}.
\end{equation}

To account for the model's inherent performance and potential noise affecting the overshadowing rate, we also introduce $R_{dom}$ for dominant knowledge inputs:

\begin{equation}
    R_{dom} = p(Y_{dom}|P_{dom}) = \frac{M_{dom}}{N_{dom}},
\end{equation}

which represents the recall rate for $P_{dom}$ query-response instances. 

The Relative Overshadowing rate ($\mathcal{RO}$) is then defined as:

\vspace{-0.3cm}
\begin{equation}
     \mathcal{RO} = \frac{\mathcal{AO}}{R_{dom}} = \frac{p(Y_{dom}|P_{sub})}{p(Y_{dom}|P_{dom})} .
\end{equation}
\vspace{-0.3cm}

\subsubsection{Overshadowing Influence Factors} 

\textbf{Knowledge Popularity (P)} is the fundamental cause of the knowledge overshadowing phenomenon and serves as its primary influencing factor. P is defined as the ratio of $N_{dom}$ to $N_{sub}$ in a group, 

\begin{equation}
    P = \frac{N_{dom}}{N_{sub}}.
\end{equation}

In the case of Figure \ref{fig:sec3case}, $ P = 2 $. A higher value of P signifies a more severe data imbalance.

\textbf{Model Size (M)}, referring to the number of model parameters, also impacts knowledge overshadowing. A larger M generally implies stronger generalization capabilities, causing the model to rapidly generalize the $X_{dom} \leftrightarrow X_{bg}$ to instances involving $X_{sub} \leftrightarrow X_{bg}$ and exacerbating overshadowing.

In addition to the factors mentioned in \cite{zhang2025law, zhang2024knowledge}, aiming to analyze the dynamic evolution of knowledge overshadowing during the training process, we extend our consideration to  total number of tokens, the \textbf{Dataset Size (D)} in the training set. Furthermore, the \textbf{average loss proportion of subordinate knowledge $\mathcal{LP}$ within an epoch during training} is defined as 

\begin{equation}
    \mathcal{LP} = \frac{loss(P_{sub})}{total \; loss},  
\end{equation}
 which measures the allocation of the model's efforts for the optimization of subordinate knowledge and recovery from knowledge overshadowing.

\subsection{Analysis~Framework~\framework}

Our proposed knowledge circuit-based overshadowing analysis framework involves the overshadowing dynamics analysis during training and circuit-based internal mechanism analysis.

\subsubsection{Overshadowing Dynamic Analysis}

Our framework provides a novel dynamic analysis of knowledge overshadowing during model training. By manipulating P, M, and D, we monitor $\mathcal{RO}$ across epochs. We focus on identifying the onset, duration, and recovery stages of overshadowing to understand their modulation by P, M, and D. Recognizing P as a key factor, we also explore how $\mathcal{LP}$ co-evolves with $\mathcal{RO}$ under these variables, aiming to uncover the role of $\mathcal{LP}$ in explaining overshadowing dynamics.

\subsubsection{Knowledge Circuit Construction}

We construct the knowledge circuit, a sparse computational subgraph $C \subseteq G$, where $G=(V,E)$ is the directed acyclic graph representation of LLMs. The node set $V$ encompasses input embeddings, attention heads, MLP layers, and output logits. The edge set $E$ represents the information flow between these components. Our goal is to identify a subgraph $C$ that is critical for recognizing the key component of given input prompt, particularly in knowledge overshadowing, is $\{X_{dom}, X_{sub}\}$, the difference between $P_{dom}$ and $P_{sub}$. The adapted construction method is similar to Edge Attribution Patching  with Integrated Gradients (EAP-IG) \cite{hanna2024have}, which involves:

\textbf{Paired inputs.} For a given $X_{bg}$, we create a pair of input prompts, the clean input $P_{sub}$ corresponding to our expected output $Y_{sub}$ in the overshadowing case, and the corrupt input $P_{dom}$ serving for the contrast mentioned below.

\textbf{Activation difference calculation.} After running this pair of inputs through the model, we calculate the difference in activation values $\Delta A(v_p), \Delta A(v_c) $ for the parent node $v_p$ and child node $v_c$ of each edge under these distinct inputs, $ \Delta A(v_p) = A_{clean}(v_p) - A_{corrupt}(v_p), $
where $A_{clean}(v_p)$ and $A_{corrupt}(v_p)$ are the activations of the parent node for the clean input $P_{sub}$ and the corrupt input $P_{dom}$, respectively.

  \textbf{Edge score calculation.} The importance $S(e)$ of an edge $e=(v_p, v_c)$ is scored based on how patching $v_p$'s activation (using $\Delta A(v_p)$) influences a metric $\mathcal{M}$, which assesses the model's ability to correctly output $Y_{sub}$ rather than $Y_{dom}$ for $P_{sub}$ inputs. Using the Integrated Gradients (IG), $S(e)$ is approximated as:
  \vspace{-0.2cm}
\begin{equation}
     S(e) \approx Exp\left[ \Delta A(v_p) \cdot \frac{\partial \mathcal{M}(Y_{sub} | P_{sub})}{\partial A(v_p)} \right] .
\end{equation}
 \vspace{-0.05cm}
 
\textbf{Circuit construction.} Edges with scores $|S(e)|$ below a threshold $\tau$ are pruned. The remaining subgraph forms the knowledge circuit $C$. See more details about construction in Appendix \ref{app:construction}

\textbf{Circuit in Knowledge Overshadowing.} In the context of knowledge overshadowing, the metric $\mathcal{M}$ is defined as the difference between the output logits of $Y_{sub}$ and $Y_{dom}$. A positive value for $\mathcal{M}$ indicates that the model correctly predicts $Y_{sub}$. Therefore, we can represent the circuit where overshadowing occurs as $C_{overshadowed} = \{V, E\}$ with $\mathcal{M} < 0$ and  $S(E|P_{sub}, Y_{dom}) \geq \tau$. Furthermore, the competition between the dominant and subordinate knowledge pairs during circuit construction is quantified by the activation values of different input $P_{dom}$ and $P_{sub}$, $\Delta A(v_p) = A_{P_{sub}}(v_p) - A_{P_{dom}}(v_p)$.

\subsubsection{Circuit-based Analysis}
\framework mainly focuses on the attention heads in $C$ and follows these steps:

\textbf{Node Attention Analysis.} We identify high-attention heads within $C$ by examining their attention scores and patterns, specifically their focus on $\{X_{dom}, X_{sub}\}$.

\textbf{Circuit Structure Analysis.} We then trace the information flow of these high-attention heads by identifying their parent and child nodes to understand their structural role. Nodes consistently retained in circuits built with different thresholds $\tau$ are also analyzed as key components.

\textbf{Layer-wise Logit Evolution.} Finally, using logit lens \cite{nostalgebraist2020interpreting}, we inspect the evolving output logits at layers associated with key nodes. This validates if their captured information contributes to the model's prediction as expected.

\subsubsection{Circuit-based Overshadowing Recovery}

Inspired by the goal of knowledge circuits to maximize sensitivity to the distinguishing features between $X_{dom}$ and $X_{sub}$, we propose a circuit-based method to alleviate overshadowing.
This involves optimizing the circuit structure by tuning the edge pruning threshold $\tau$ to obtain an optimal circuit, $C_{opt}$.
The optimization is formulated as:
\vspace{-0.2cm}
\begin{equation}
    \tau_{opt} = \operatorname*{arg\,max}_{\tau} \mathcal{M} (C_{opt}(\tau), P_{sub}, Y_{sub}), 
    \label{eq:circuit_optimization}
\end{equation}

where $\mathcal{M}$ measures the circuit's ability to distinguish $\{X_{dom}, X_{sub}\}$. We use a two-stage optimization method to find the optimal number of edges. First, we build the circuit by adding edges at regular intervals, revealing that the circuit's performance exhibits a multi-modal relationship with the edge count. Subsequently, we apply the golden-section search algorithm within the optimal range to determine the $\tau_{opt}$ for the $C_{opt}$. This  results in a $C_{opt}(\tau_{opt})$ which is expected to mitigate or eliminate the overshadowing effect for specific input prompts.

To automate the overshadowing recovery process and extend its applicability, we simplify the relative pointwise mutual information (R-PMI) method from \cite{zhang2025law, zhang2024knowledge} to identify  $X_{sub}$ within $P_{sub}$. 
First, we obtain the top-$k$ next-token candidates, $V_{top}(P_{sub})$, by feeding $P_{sub}$ to the model. 
Then, we iteratively generate contrastive prompts $P'_{sub}$ by masking (in our implementation, by deleting) each token $X'_{sub}$ from $P_{sub}$. For each $P'_{sub}$, we acquire its top-$k$ candidates $V_{top}(P'_{sub})$. 
The R-PMI for each token $y_i$ in the intersection $V_{top}(P_{sub}) \cap V_{top}(P'_{sub})$ is calculated as: 
\vspace{-0.2cm}
\begin{equation}
    R\textnormal{-}PMI_i = \log \frac{p(y_i | P_{sub})}{ p(y_i | P'_{sub})}.
\end{equation}
 Then, we sum the negative R-PMI values to get $S_{R\textnormal{-}PMI} = \sum \min(R\textnormal{-}PMI_i, 0)$. The $X'_{sub}$ yielding the minimum $S_{R\textnormal{-}PMI}$ is identified as $X_{sub}$.

Furthermore, the $Y_{sub}$ is determined as the token $y_i$ from $V_{top}(P_{sub})$ whose average rank improves most significantly when non-subordinate components $X'_{sub}$, often the $X_{bg}$, are masked. $Y_{dom}$ is identified as $y_i$ that has the highest average rank across all $P'_{sub}$. For circuit construction, $X_{dom}$ within $P_{dom}$ is replaced by a generic placeholder like "something", as shown in Figure \ref{fig:case}. Combining this streamlined approach enables broader application of our knowledge circuit-based overshadowing recovery. See more details in Appendix \ref{app:location}

% Inspired by the goal of knowledge circuits to maximize sensitivity to $\{X_{dom}, X_{sub}\}$, we explore a potential circuit-based method to alleviate overshadowing, which involves optimizing the number of edges  to obtain an optimal circuit $C_{opt}$ and then reduce the overshadowing effect.

% This is formulated as a single-objective optimization problem. The optimization variable is the edge pruning threshold $\tau$. The optimal $\tau$ that results in maximum overall $\mathcal{M}$ helps to construct $C_{opt}$ that more focus on $\{X_{dom}, X_{sub}\}$, consequently alleviating or eliminating the overshadowing effect on specific input prompt.

\begin{figure*}
  \centering
  \begin{subfigure}[b]{0.32\linewidth}
    \includegraphics[width=\linewidth]{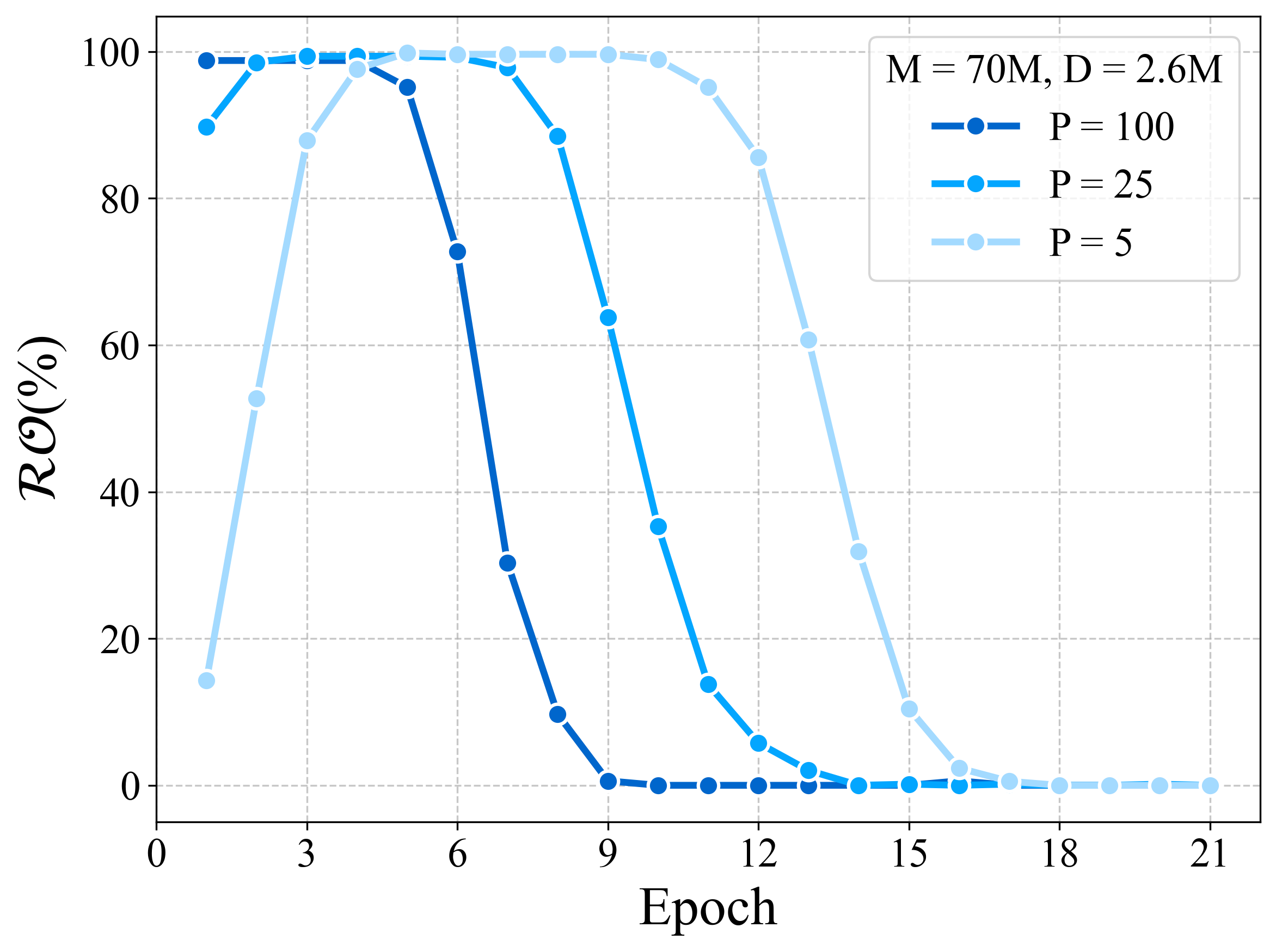} % 替换为你的图片路径
    \caption{Popularity (P)}
    \label{fig:ro_p}
  \end{subfigure}
  \hfill % 子图间距
  \begin{subfigure}[b]{0.32\linewidth}
    \includegraphics[width=\linewidth]{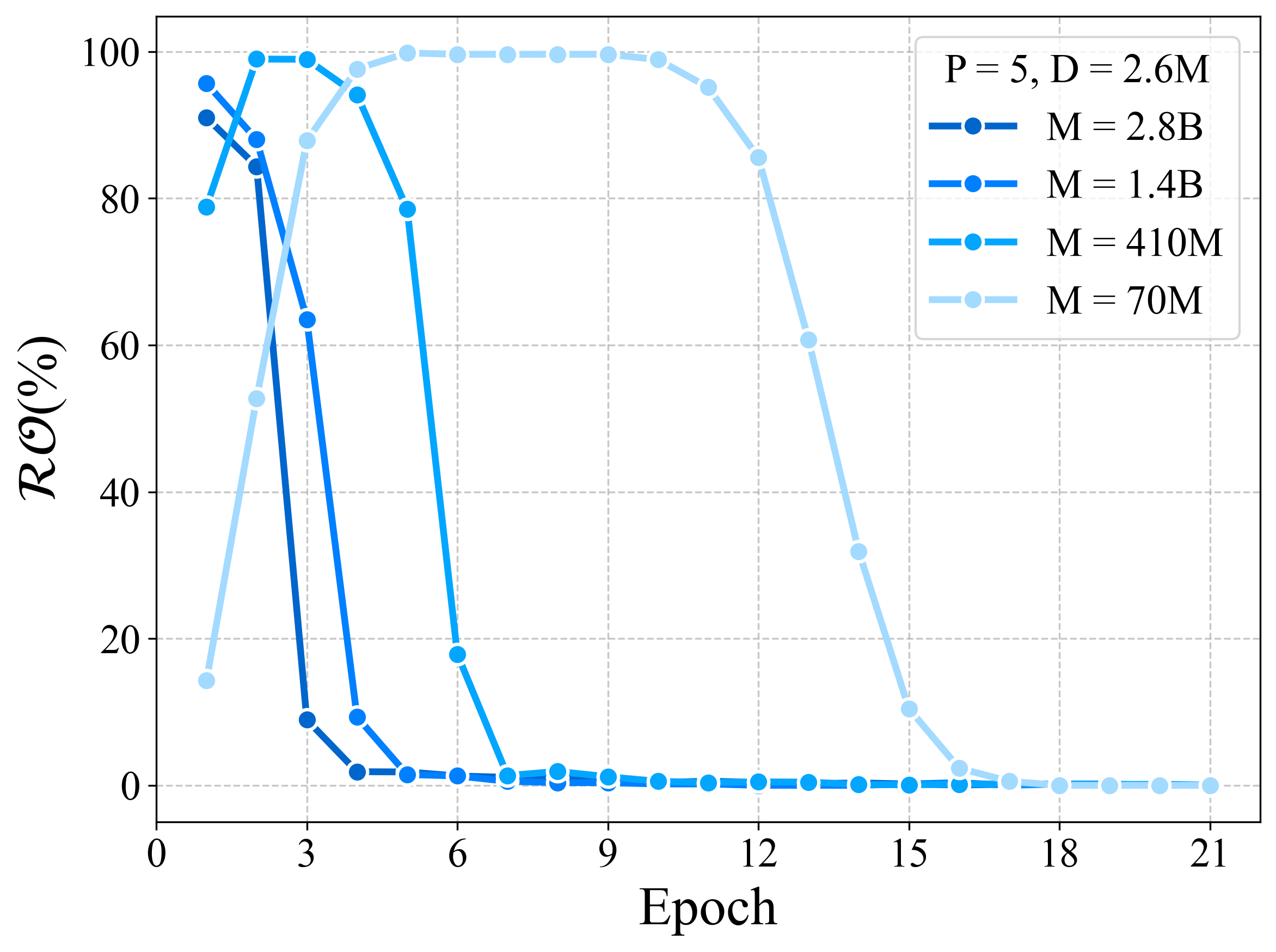} % 替换为你的图片路径
    \caption{Model Size (M)}
    \label{fig:ro_m}
  \end{subfigure}
  \hfill % 子图间距
  \begin{subfigure}[b]{0.32\linewidth}
    \includegraphics[width=\linewidth]{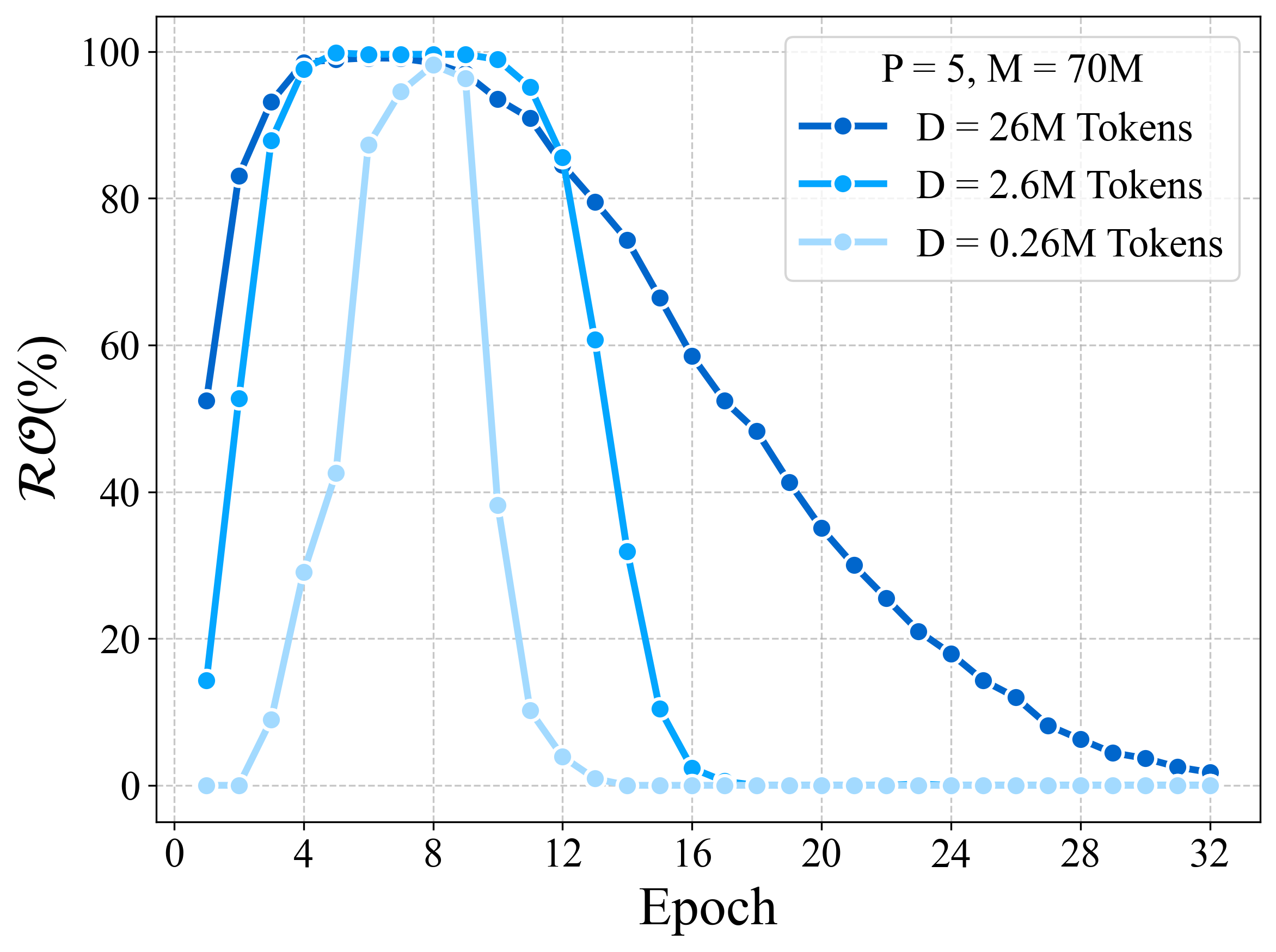} % 替换为你的图片路径
    \caption{Dataset Size (D)}
    \label{fig:ro_d}
  \end{subfigure}
  \begin{subfigure}[b]{0.32\linewidth}
    \includegraphics[width=\linewidth]{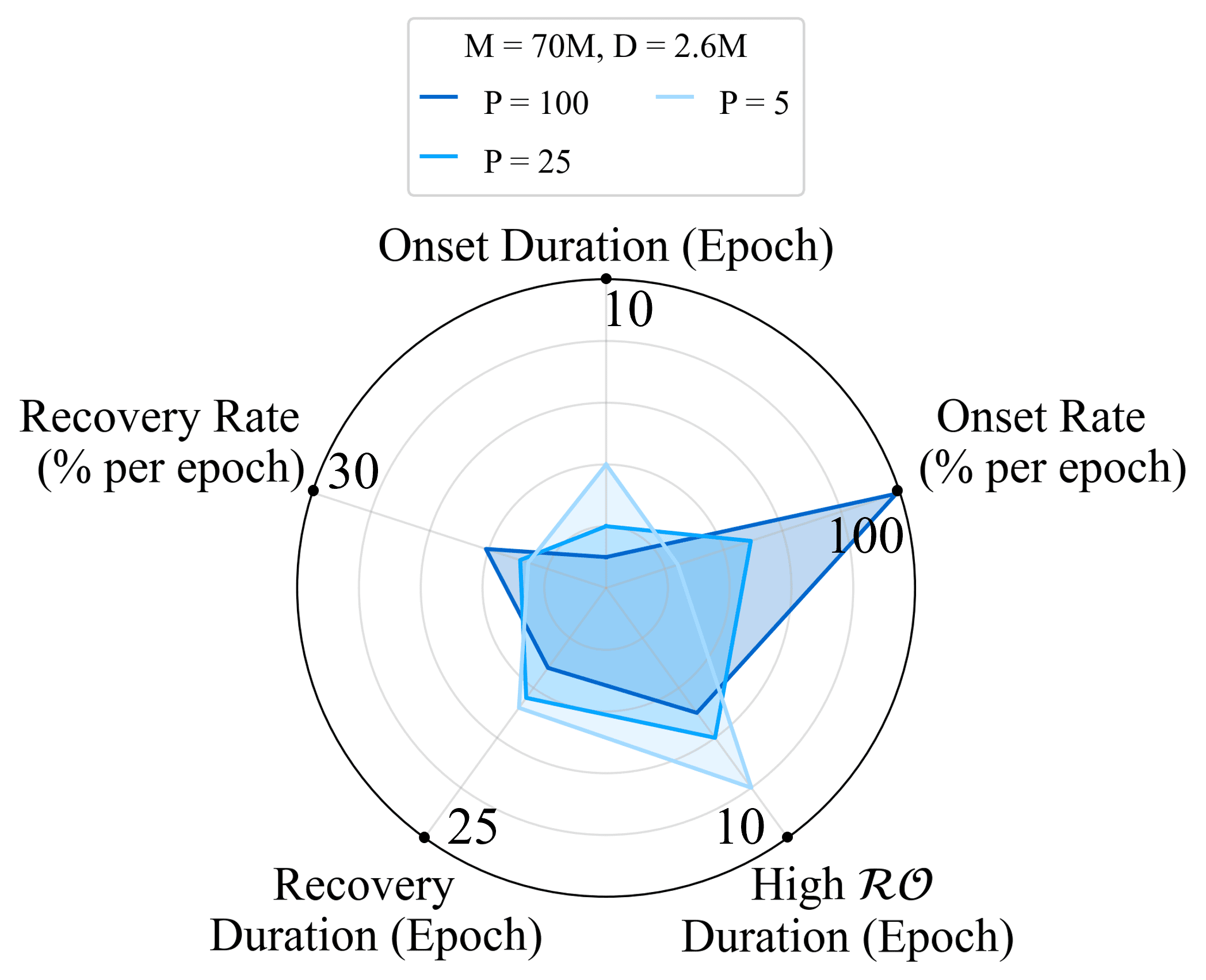} % 替换为你的图片路径
    \caption{Popularity (P)}
    \label{fig:radar_p}
  \end{subfigure}
  \begin{subfigure}[b]{0.32\linewidth}
    \includegraphics[width=\linewidth]{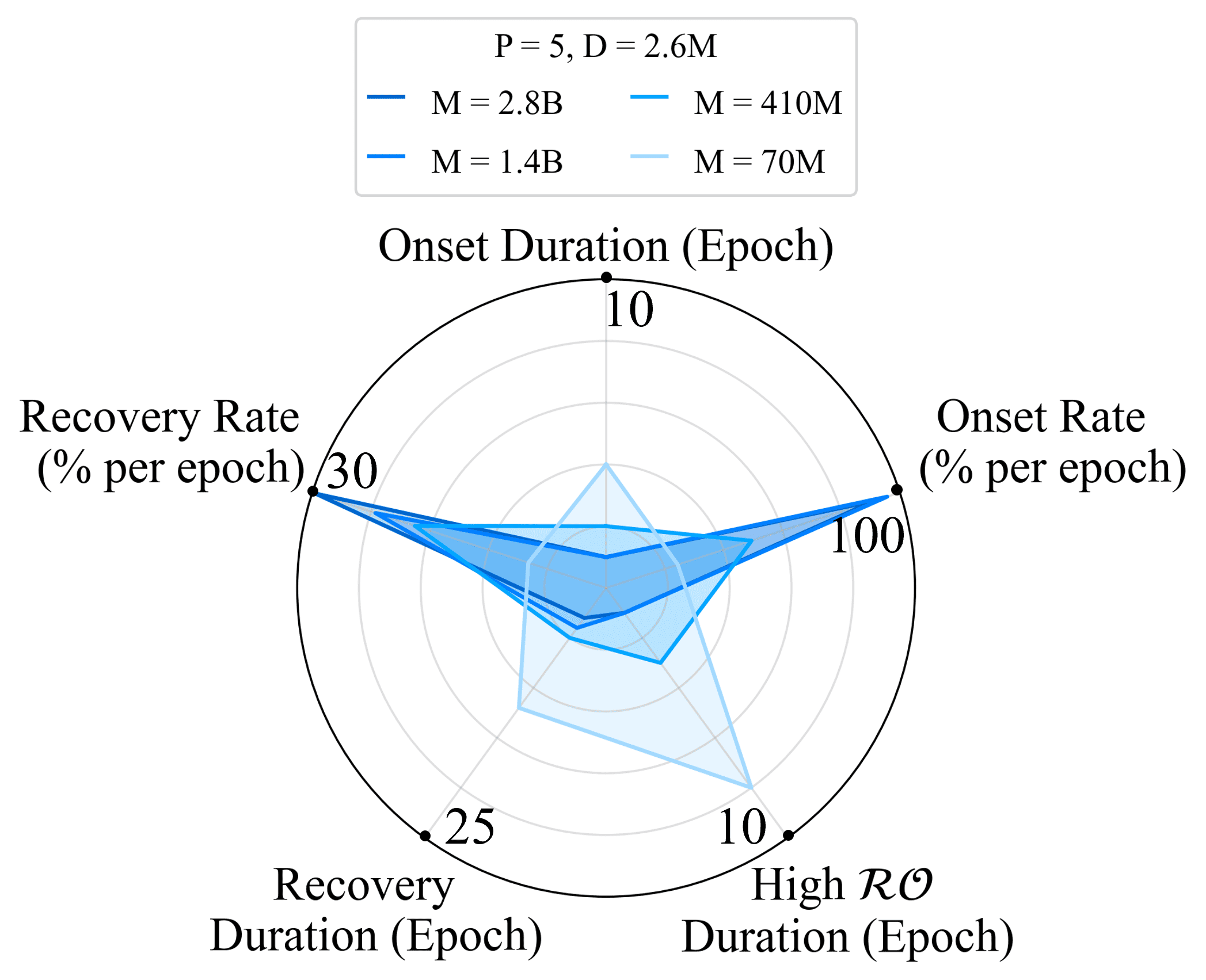} % 替换为你的图片路径
    \caption{Model Size (M)}
    \label{fig:radar_m}
  \end{subfigure}
  \begin{subfigure}[b]{0.32\linewidth}
    \includegraphics[width=\linewidth]{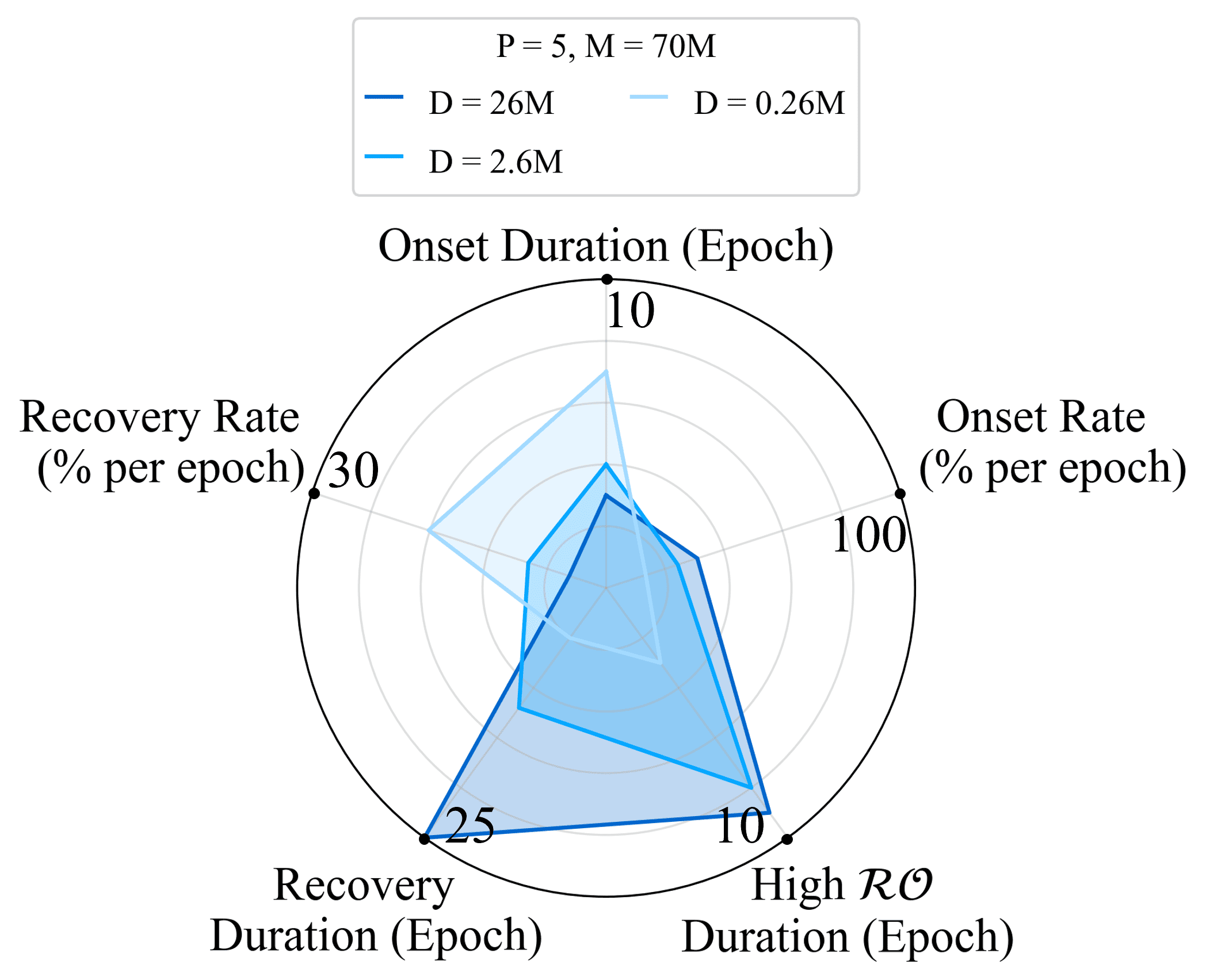} % 替换为你的图片路径
    \caption{Dataset Size (D)}
    \label{fig:radar_d}
  \end{subfigure}
  
 \vspace{-0.1cm}
   \caption{(a) $\sim$ (c) show the dynamic variation of $\mathcal{RO}$ relating to P, M and D in model training phase. Higher Knowledge Popularity (P) and Model Size (M) tend to result in an earlier onset, shorter duration, and quicker recovery from knowledge overshadowing. In contrast, a larger Dataset Size (D) also leads to an earlier onset but is associated with a slower recovery phase. (d) $\sim$ (f) show the duration of onset stage, high  $\mathcal{RO}$ (> 90\%) stage as well as recovery stage, and $\mathcal{RO}$'s rate of change during the onset and recovery stages.
}
     \label{fig:pmd}
   \vspace{-0.3cm}
\end{figure*}
\vspace{-0.2cm}
\section{Experiment}

\subsection{Experiment Setup}

\begin{figure*}[h]
  \centering
  \begin{subfigure}[b]{0.32\linewidth}
    \includegraphics[width=\linewidth]{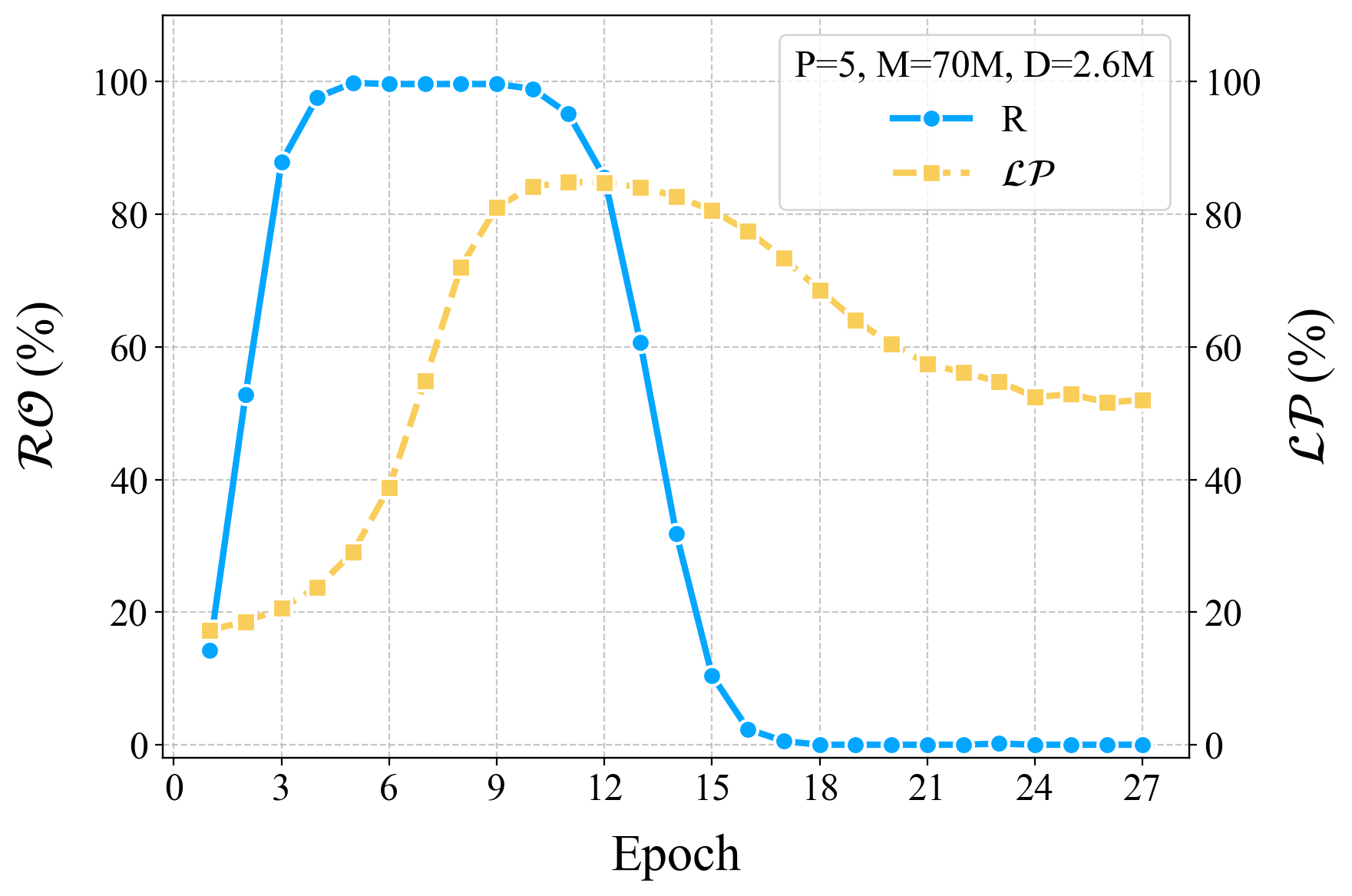} % 替换为你的图片路径
    \caption{M = 70M}
    \label{fig:ro_lp_70m}
  \end{subfigure}
  \hfill % 子图间距
  \begin{subfigure}[b]{0.32\linewidth}
    \includegraphics[width=\linewidth]{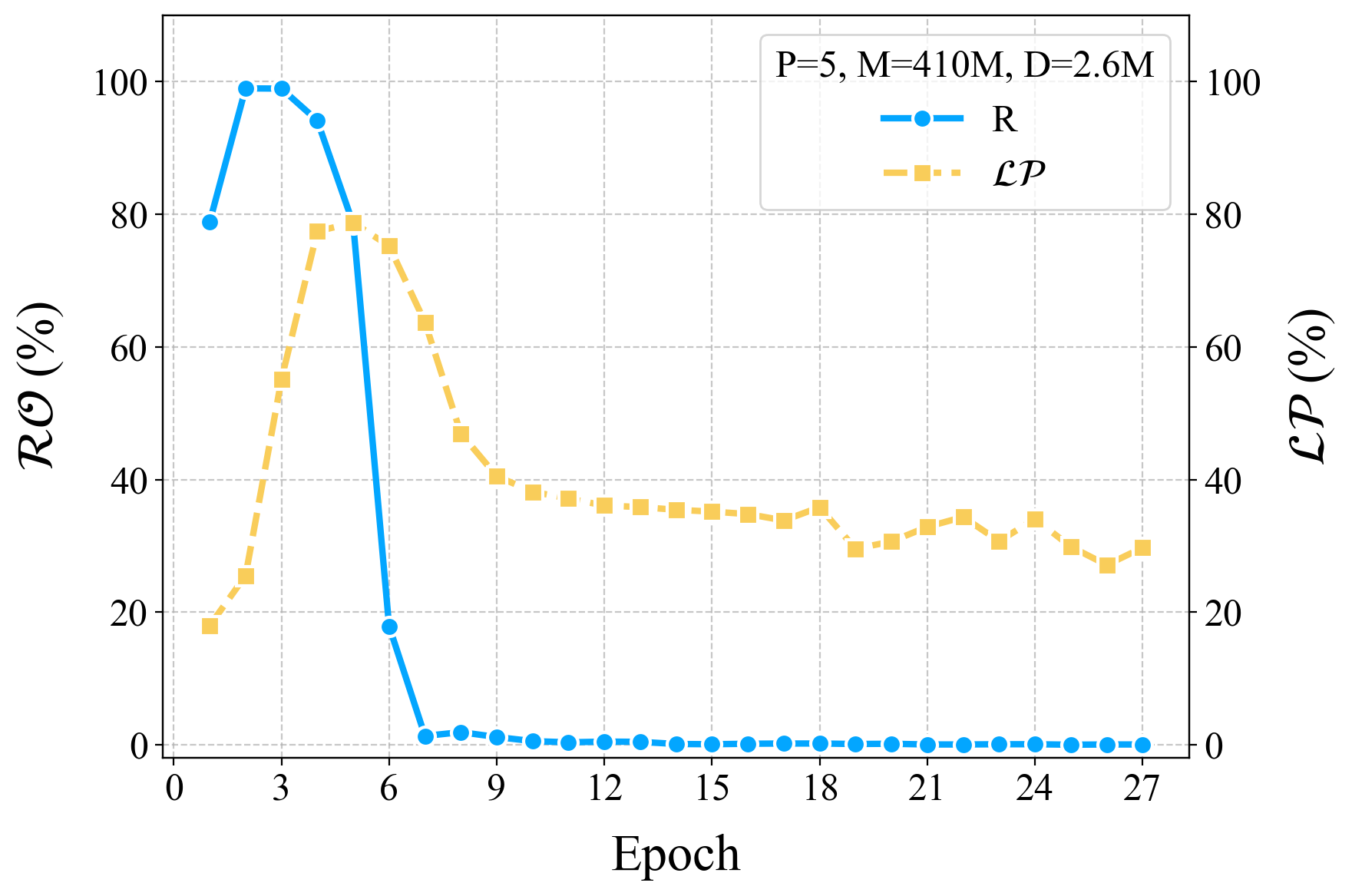} % 替换为你的图片路径
    \caption{M = 410M}
    \label{fig:ro_lp_410m}
  \end{subfigure}
  \hfill % 子图间距
  \begin{subfigure}[b]{0.32\linewidth}
    \includegraphics[width=\linewidth]{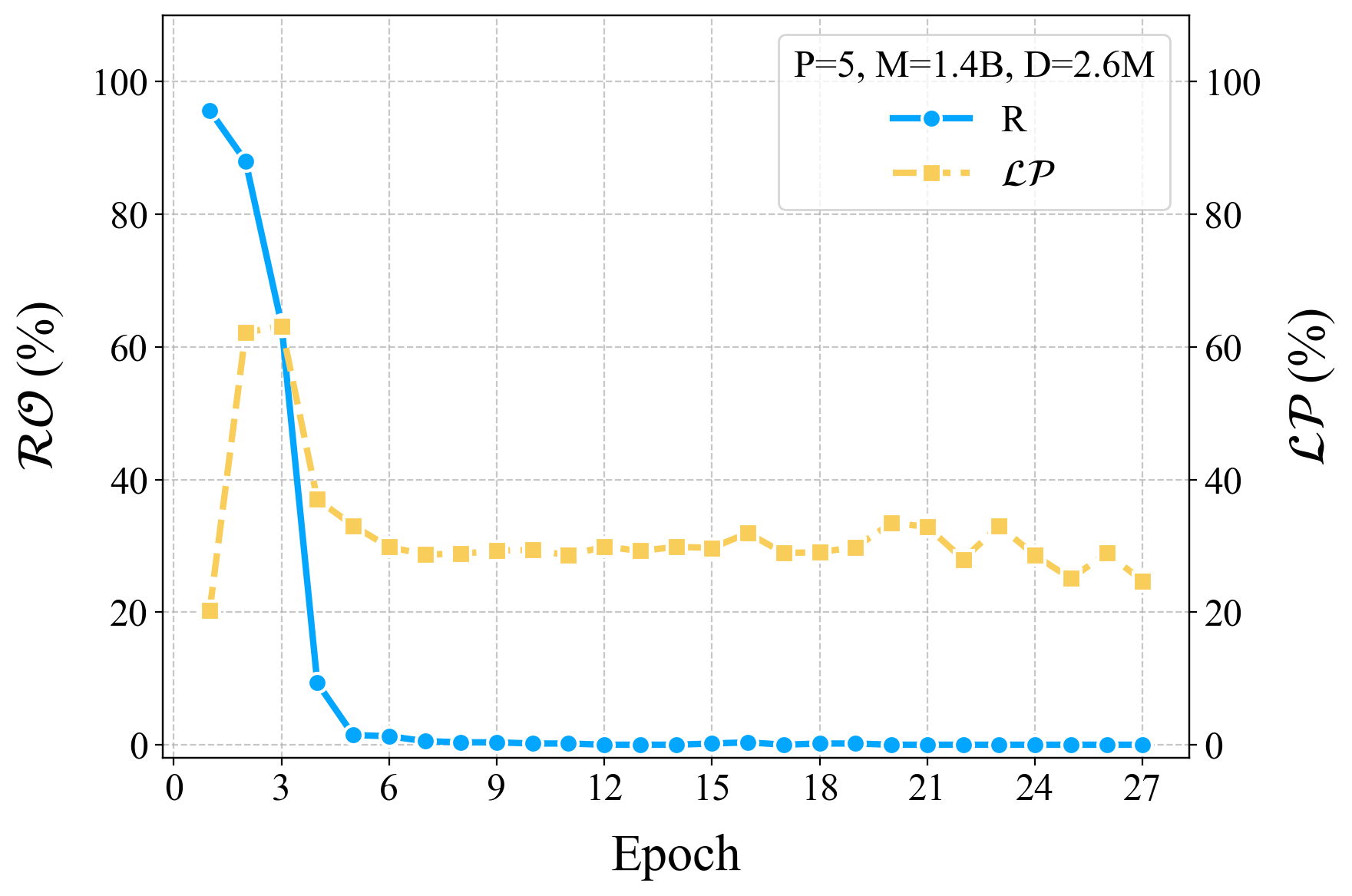} % 替换为你的图片路径
    \caption{M = 1.4B}
    \label{fig:ro_lp_1.4b}
    
  \end{subfigure}
\vspace{-0.2cm}
  \caption{The co-evolution between $\mathcal{RO}$ \& $\mathcal{LP}$ in different M. Early high overall loss and low $\mathcal{LP}$ leads to intensive optimization of $P_{dom}$ and $\mathcal{RO}$ rises up to near 100\%. As training progresses, subordinate knowledge loss proportion ($\mathcal{LP}$) rises, shifting optimization focus to $P_{sub}$ errors, initiating $\mathcal{RO}$'s recovery phase, validated across models.}

  \label{fig:lp}
    \vspace{-0.5cm}
\end{figure*}

\subsubsection{Dataset}

\textbf{Synthetic dataset.} To investigate the dynamic characteristics and influencing factors of the $\mathcal{RO}$ during training under controlled conditions,and to minimize the complexities and semantic relationships inherent in natural language, we construct a synthetic dataset and train models from scratch. 

Follow \cite{zhang2025law}, we first fix the length of $X_{bg}$ at 4 tokens and the lengths of $X_{dom}, X_{sub}, Y_{dom}$, $Y_{sub}$ at 1. For the dataset size (D), we experiment with 0.26  (D = 0.26M), 2.6 (D = 2.6M) and 26 million tokens (D = 26M). For popularity (P), we set P values as 5, 25, and 100.

Then, for a specific combination of P and D, there are several distinct groups to achieve the target D. Each group comprises P distinct $X_{dom}$ and a single $Y_{dom}$ for dominant knowledge, one $X_{sub}$ and $Y_{sub}$ for subordinate knowledge. The $X_{bg}$ is shared. Thus, there are P unique $\{P_{dom}, Y_{dom}\}$ and one $\{P_{sub}, Y_{sub}\}$ in a group. All the tokens are randomly sampled. See more details of our dataset in Appendix \ref{app:dataset}.

\textbf{fine-tuning Dataset}.
We construct a fine-tuning dataset to evaluate circuit-base overshadowing recovery method. Utilizing virtual knowledge, we preserve the natural language semantics while avoiding prior knowledge editing that could interfere with the natural occurrence of overshadowing. For this dataset, we set P = 5 and D = 1M. The Appendix \ref{app:dynamics} shows the dynamics analysis based on fine-tuning dataset. 

\subsubsection{Models Evaluated}
We employ models from the Pythia suite \cite{biderman2023pythia}, specifically: Pythia-70M, Pythia-410M, Pythia-1.4B, and Pythia-2.8B, corresponding to model sizes (M) of M = 70M, 410M, 1.4B and 2.8B, respectively. Tokens randomly sampled to build synthetic dataset is from Pythia tokenizer.

\subsubsection{Training}
A uniform learning rate of $10^{-5}$ and batch size of 16 are used for both dataset. Training is conducted on NVIDIA A800 GPUs.
\vspace{-0.2cm}

\begin{figure*}[t]
    \centering
    \includegraphics[width=\textwidth]{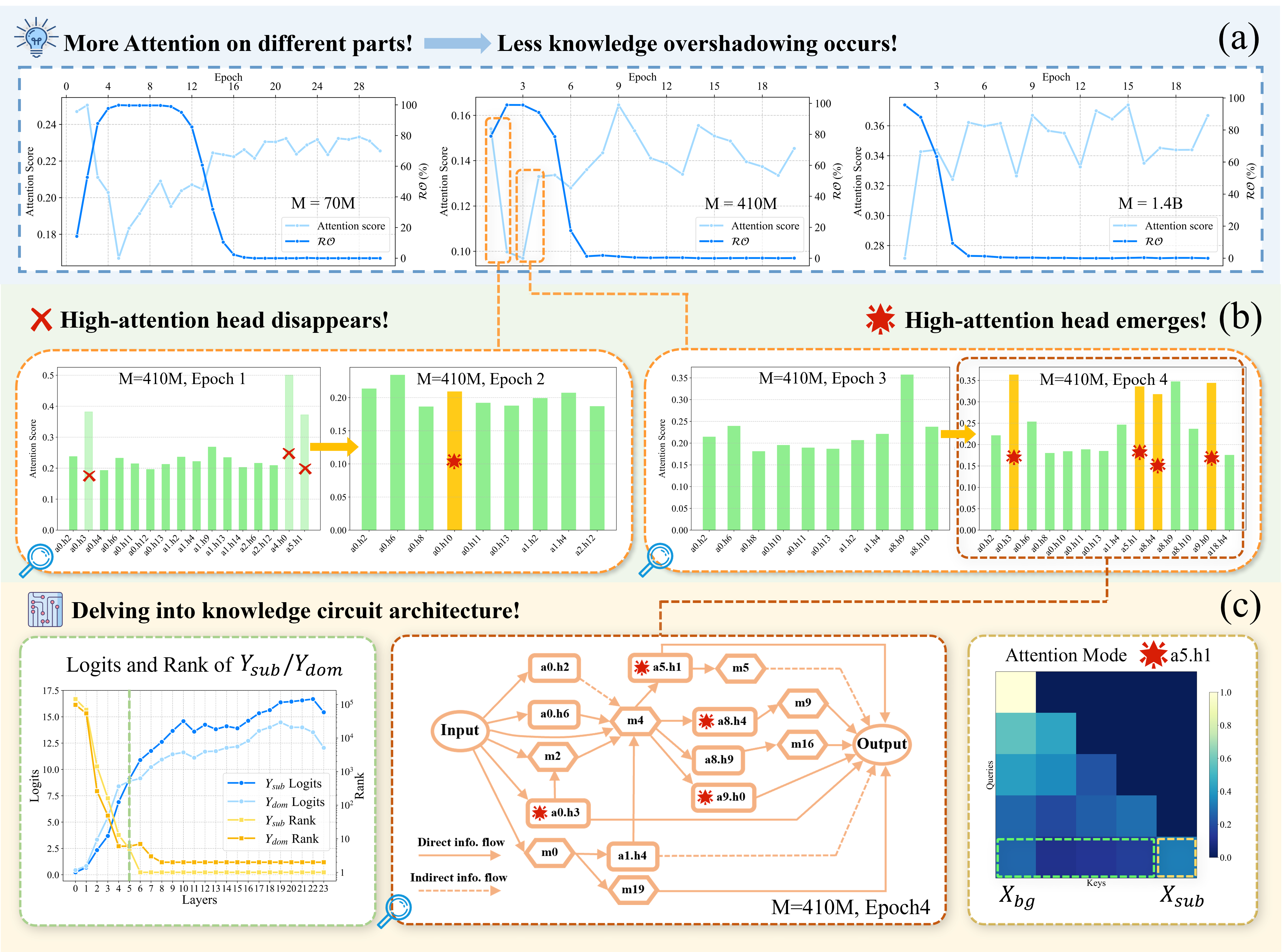}
    \caption{The circuit-based analysis of proposed \framework. (a) shows the average attention scores allocated to $\{ X_{dom}, X_{sub} \}$ across epoch. (b) focuses on the onset and recovery phases and shows that the appear and disappear of high-attention heads (attention score greater than 0.2) attribute to the reduce and raise of $\mathcal{RO}$. (C) shows the logits and ranks of $Y_{sub}, Y_{dom}$ across layers. The main structure of circuit (with 400 edges totally)  and attention mode show that high-attention head a5.h1 and MLP layer m5 contribute to the juncture of $Y_{sub}$ 's rank. Solid lines denote direct information flow, while dashed lines indicate indirect flow in circuit structure map.}

    \label{fig:find3}
    \vspace{-0.5cm}
\end{figure*}

\subsubsection{Evaluation}
To measure $\mathcal{RO}$, we randomly sample 500 $P_{dom}$ and 500 $P_{sub}$ prompts for evaluation after each training epoch. The $\mathcal{LP}$ is recorded within each epoch. The results are shown in Figures \ref{fig:pmd} and \ref{fig:lp}.

\subsection{Main Result}

Based on the experiments described above, using the circuit to analyze and optimize overshadowing, our investigation yields the following findings.

\textbf{A higher value of P and M can lead to an earlier onset, shorter duration, and quicker recovery of the knowledge overshadowing. Distinctly, a larger D contributes to the earlier onset but also a slower recovery from overshadowing.}

As shown in Figures \ref{fig:ro_p} and \ref{fig:radar_p} (M=70M, D=2.6M), increasing P (from 5 to 100) significantly shortens or even eliminates the onset phase. This is attributed to more prominent dominant patterns $X_{dom} \leftrightarrow X_{bg}$ being learned and generalized rapidly, even within the first epoch. The duration phase also decreases with higher P,  because a larger P and a fixed D implies fewer groups of knowledge pairs, leading to less diversity of $P_{sub}$,
allowing the model to learn all overshadowed $P_{sub}$ and recover from overshadowing more quickly.

Figures \ref{fig:ro_m} and \ref{fig:radar_m}  (P=5, D=2.6M) demonstrate that larger models (M) exhibit shorter or absent onset phases, indicating an earlier occurrence of knowledge overshadowing. This is due to the stronger generalization capabilities of larger models, leading them to quickly learn and overgeneralize dominant patterns. However, larger models also show a shortened duration phase and a rapid decline in $\mathcal{RO}$ during recovery, suggesting enhanced capacity to differentiate $\{X_{dom}, X_{sub}\}$, thus recovering faster despite earlier overshadowing.

\begin{figure*}[t]
    \centering
    \includegraphics[width=\textwidth]{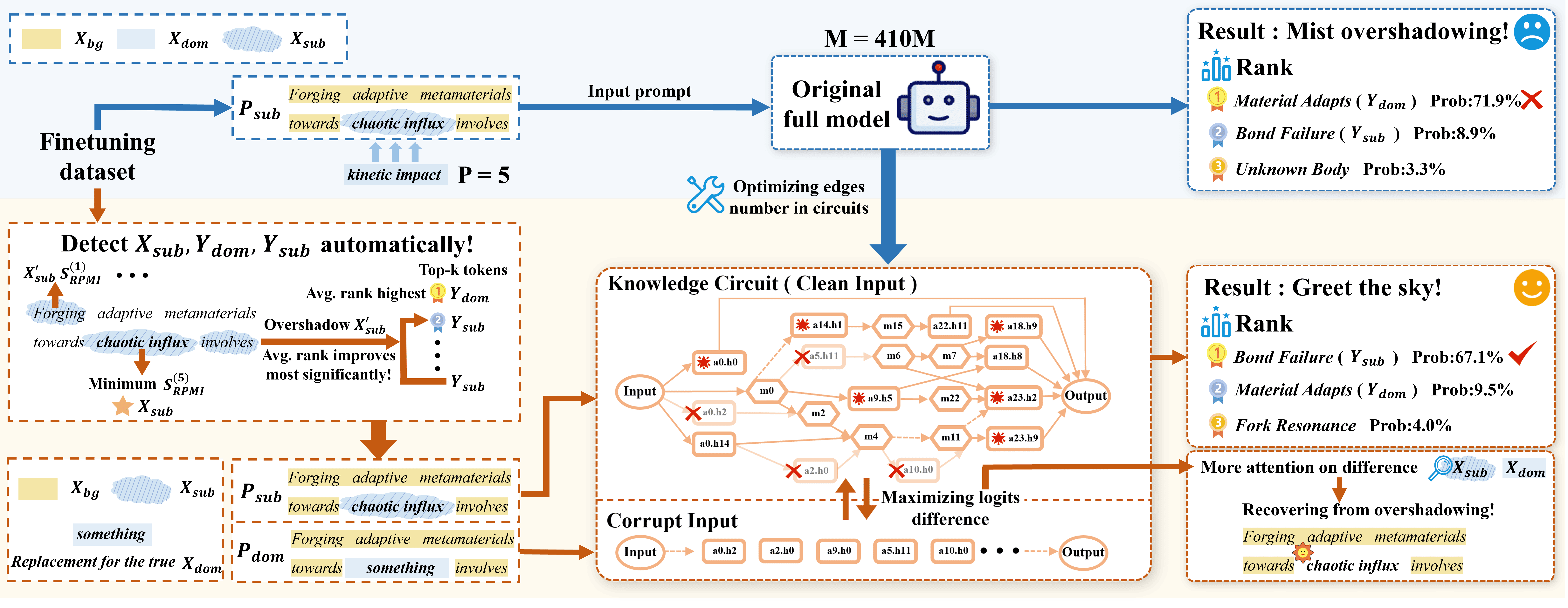}
    \caption{Overshadowing recovery via optimized circuit. In the fine-tuned model, $ X_{dom} \leftrightarrow X_{bg}$ causes the original full model to incorrectly predict $Y_{dom}$. First, we detect the $X_{sub}, Y_{sub}, Y_{dom}$ automatically by calculating the minimum $S_{R\textnormal{-}PMI}$ . Then, optimizing the knowledge circuit by pruning edges to keep only key nodes enhances the attention on $\{X_{dom}, X_{sub}\}$, enabling  the recovery from overshadowing and facilitating correct $Y_{sub}$ prediction.}
    \label{fig:case}
    \vspace{-0.5cm}
\end{figure*}

\vspace{-0.1cm}

In Figures \ref{fig:ro_d} and \ref{fig:radar_d} (P=5, M=70M) larger D leads to an earlier onset of overshadowing, with $\mathcal{RO}$  peaking sooner. This is because more data per epoch provides more iterations and exposure to the dominant pattern, accelerating its generalization. Conversely, the recovery phase is prolonged with larger datasets. The increased diversity of $P_{sub}$ in larger datasets requires more epochs for the model to learn all instances and recover.

Notably, across various parameter combinations, $\mathcal{RO}$ often approaches 100\% in the early training stages and finally recovers to 0\%. We hypothesize this phenomenon stems from initial high-loss state, where optimization efforts disproportionately focus on reducing the larger loss contribution from $P_{dom}$. Our subsequent observations regarding $\mathcal{LP}$ corroborate this.

\textbf{The dynamic nature of knowledge overshadowing arises from the co-evolution relationship between the $\mathcal{LP}$ and $\mathcal{RO}$.} As depicted in Figure \ref{fig:lp}, in the early training stages, when the overall loss is high and the contribution from $P_{sub}$ is small, the optimization process tends to concentrate on the $P_{dom}$ and $\mathcal{RO}$ rapidly approaches 100\%.
\vspace{-0.1cm}
However, as training progresses and $\mathcal{LP}$ begins to rise, eventually nearing its peak, $P_{sub}$ takes a substantial portion of the remaining loss. At this juncture, the model's optimization efforts shift to focus on these errors from $P_{sub}$. This shift initiates the recovery phase, leading to a decline in $\mathcal{RO}$. Therefore, the insufficient optimization results in the overshadowing, which is consistent across varying M.

 % Triggered by the initial high $\mathcal{RO}$, the $\mathcal{LP}$ serves as a critical driver for following hallucination optimization. In the initial training phase, the absence of $\mathcal{LP}$
 %  contributes to elevated knowledge overshadowing rates, which almost reaches 100\%. Subsequently, the rapid rise of $\mathcal{LP}$ facilitates the mitigation of overshadowing effects, leading to the decline in those overshadowing incidents.

 \textbf{The occurrence of overshadowing in pre-trained LLMs can be understood with dynamics analysis results above.} Initially, large M and D promote the rapid generalization of $X_{dom} \leftrightarrow X_{bg}$ and overshadowing onset. The subsequent inadequate optimization of the $P_{sub}$ is exacerbated by the sheer scale and diversity of training data, which brings prolonged overshadowing effect and exceeded sharp recovery effect from large M (e.g., trillion tokens for training Llama-2-7B).

\vspace{-0.1cm}

\textbf{The knowledge circuit's attentional allocation to differences between dominant and overshadowed knowledge inputs influences the extent of knowledge overshadowing.} Figure \ref{fig:find3} (a) shows the variation of circuit's average attention scores on $\{X_{dom}, X_{sub}\}$ throughout the training phase. The higher average attention alleviates overshadowing, while lower attention exacerbates it.

\vspace{-0.1cm}

Figure \ref{fig:find3} (b) shows the circuit dynamics across the onset and recovery phases of knowledge overshadowing. These bar plots show the attention scores of individual attention heads in a specific epoch, indicating that when the $\mathcal{RO}$ declines, some attention heads, defined as high-attention heads, exhibiting high focus on $\{X_{dom}, X_{sub}\}$ emerge within the circuit. The threshold for high-attention is set at 0.2 for the length of $X_{sub}$ is one fifth of the length of input prompt in synthetic dataset.
% These attention heads are defined as high-attention heads when the attention scores on $\{X_{dom}, X_{sub}\}$ go beyond 0.2, this is because the length of $X_{sub}$ is one fifth of the length of input prompt in synthetic dataset. 
Conversely, when knowledge overshadowing intensifies, a subset of these critical attention heads tends to disappear from the circuit. 

\vspace{-0.12cm}

%%%%%%%%%%%%%%%%%%%%%%%%%%%%%%%%%%%%%%%%%%%%%%%%%%%%%%%%%%%%%5

\begin{table*}[t]
\caption{Ablation Study for high-attention Heads on fine-tuning Dataset. The model's performance degrades with the ablation of high-attention heads, which validates the circuit's faithfulness.}
\begin{tblr}{ % <--- 注意这里的括号
  width = \linewidth,
  colspec = {Q[c,m,wd=0.15\linewidth]   % Model
             Q[c,m,wd=0.4\linewidth]   % Metrics
             Q[c,m,wd=0.115\linewidth]   % 10% (改为居中 c,m)
             Q[c,m,wd=0.115\linewidth]   % 20% (改为居中 c,m)
             Q[c,m,wd=0.115\linewidth]}, % 50% (改为居中 c,m)
  % --- 字体设置 ---
  cells = {font=\small},
  % --- 特定行/单元格的覆盖 ---
  row{1} = {font+=\bfseries, bg=white},
  % --- 背景色 ---
  row{2-3} = {bg=white},
  row{4-5} = {bg=white},
  % --- 线条设置 ---
  hline{1,Z} = {0.8pt,solid},
  hline{2} = {0.4pt,solid},
  hline{4} = {0.4pt,solid},
} % <--- 确保这个括号存在且位置正确
    % 表头
    \SetCell[c=1]{c,m} Model
      & \SetCell[c=1]{c,m} Proportion of Ablated High Attn. Heads
      & \SetCell[c=1]{c,m} 10\%
      & \SetCell[c=1]{c,m} 20\%
      & \SetCell[c=1]{c,m} 50\% \\
    % --- Group 1 ---
    \SetCell[r=2]{c,m} gpt2-medium
     & Performance Degradation
      & 1.03 & 1.43 & 1.67 \\
   
       & Attn. Score on $X_{sub}$ Degradation & 0.044 & 0.086 & 0.141 \\
  % --- End of Group 1 ---
    % --- Group 2 ---
     \SetCell[r=2]{c,m} pythia-410m
     & Performance Degradation
      & 0.92 & 1.30 & 1.42 \\
    % --- 下一行从第二列开始 ---
       & Attn. Score on $X_{sub}$ Degradation & 0.039 & 0.074 & 0.126 \\
\end{tblr}
\label{tab:ablation}
\end{table*}

%%%%%%%%%%%%%%%%%%%%%%%%%%%%%%%%%%%

Furthermore, by focusing on the circuit's internal mechanisms and structure within a specific epoch, as shown in Figure \ref{fig:find3} (c), we leverage the circuit-based analysis of our proposed \framework. First, according to the logits and ranks of $Y_{dom}$ and $Y_{sub}$ across layers, the 5th layer is a juncture where the rank of $Y_{sub}$ exceeds $Y_{dom}$. Subsequently, by focusing on the circuit structure, we identify the internal mechanisms driving this juncture. A high-attention head, a5.h1 (layer 5), crucially channels information to the subsequent MLP layer, m5. The a5.h1 also appears to inherit its high-attention state from earlier layers, mediated by the layer 4 MLP (m4), leading it to not only continue elevating the logits for $Y_{sub}$ but also to attenuate the previously rapid growth of $Y_{dom}$'s logits, thereby facilitating the observed rank reversal. The attention map on the right confirms its high focus on $\{X_{dom}, X_{sub}\}$.

\textbf{Knowledge circuit-guided optimization represents a promising strategy to mitigate the knowledge overshadowing effect.} Figure \ref{fig:case} illustrates our findings. 
We first fine-tune the model on fine-tuning dataset. During the recovery phase, we randomly choose some $P_{sub}$. Feeding $P_{sub}$ to the original full model (M = 410M), the prediction is still $Y_{dom}$ instead of expected $Y_{sub}$, and shows a significant gap in probability as well.

% Our investigation begins with a model fine-tuned on our virtual knowledge dataset. We select an epoch that falls within the recovery phase of knowledge overshadowing, where the overshadowing effect is declining but remains significant. Within this chosen epoch, we randomly sample pairs of $P_{dom}$ and $P_{sub}$. We first evaluate the performance of the original full model (M = 410M) on these sampled prompts. When $P_{dom}$ is is input, the prediction is still $Y_{dom}$ instead of expected $Y_{sub}$, and shows significant gap in probability as well. As mentioned in \cite{yao2025cake, yao2024knowledge}, such full models often contain numerous redundant structural components or pathways, which can introduce noise and irrelevant information, potentially contributing to knowledge overshadowing.

We then detect the $Y_{sub}, Y_{dom}$ and $X_{sub}$ for chosen $P_{sub}$ automatically, replace the $X_{sub}$ of placeholder token "something". With these components, we obtain the $P_{dom}$ as corrupt input and $ P_{sub}$ as clean input to construct a knowledge circuit. The optimized circuit graph shows that some attention heads are pruned, which are often low attention heads, or exhibit no significant attentional pattern towards the core task-relevant information. Retained high-attention heads are key to differentiating $P_{dom}$ from $P_{sub}$ which give significant attention to $\{X_{dom}, X_{sub}\}$. Some low attention heads also remain, implying that even in the circuit, processing background knowledge $X_{bg}$ and linking it to the distinctive elements $X_{sub}$ is crucial for correct inference. The performance of the optimized circuit $C_{opt}$ is then evaluated by feeding it the clean input $P_{sub}$, while $P_{dom}$ serves as the baseline for contrast. Finally, the circuit successfully produces $Y_{sub}$, demonstrating the elimination of the overshadowing effect.
%This result highlights the promise of using optimized knowledge circuits to address knowledge overshadowing.

% Future work will enhance the circuit optimization metric $\mathcal{M}$ by incorporating $Y_{sub}$'s absolute logit alongside the logit difference with $Y_{dom}$ for more effective guidance. Developing a comprehensive evaluation framework for circuit-based recovery is also crucial. These steps will evolve \framework\ into an integrated platform for efficient analysis and robust optimization of knowledge overshadowing.

\subsection{Circuit Faithfulness Study}

Based on the work of \cite{marks2024sparse, hanna2024have}, the EAP-IG method builds circuits with high faithfulness and efficiency, ensuring our findings based on circuit analysis are reliable. To verify this key property in the context of knowledge overshadowing, we conduct an ablation study on high-attention heads.

 We first construct optimal circuits for pythia-410m and gpt2-medium models trained on the fine-tuning dataset. Subsequently, we removed the top 10\%, 20\%, and 50\% of attention heads by score and observed the impact on performance metric $\mathcal{M}$ and attention scores, with results shown in Table \ref{tab:ablation}.

The results show that ablating just 10\% of the high-attention heads leads to a significant drop in both the attention scores on ${X_{dom}, X_{sub}}$ and the circuit performance metric $\mathcal{M}$. This performance degradation is substantial enough to alter the output in many cases, preventing the model from accurately predicting $Y_{sub}$. In contrast, at 50\% ablation, the drop in $\mathcal{M}$ becomes less pronounced because the model’s ability to predict both the $Y_{sub}$ and $Y_{dom}$ severely declines. These findings validate the critical impact of the nodes in optimal circuits, especially high-attention heads, on model performance and confirm the circuit's faithfulness.

\section{Conclusion}

This paper investigates hallucinations in LLMs caused by knowledge overshadowing, and introduces \framework, a novel knowledge circuit-based analysis framework.
\framework first analyzes the training dynamics of overshadowing, finding that dominant knowledge popularity, model size, and dataset size critically shape the onset, duration, and recovery of overshadowing. Apart from that, the persistent overshadowing in pre-trained models stems from inadequately optimized subordinate knowledge loss.
By analyzing knowledge circuits, we find that changes in critical attention heads' focus on subordinate knowledge directly correlate with the recovery or onset of overshadowing. Finally, optimizing these knowledge circuits presents a promising strategy for mitigating knowledge overshadowing.

\clearpage
\section*{Limitations}

Despite the insights provided by \framework, this study has several limitations that open avenues for future research:

\begin{enumerate}

    \item  Despite of the experiments on gpt2-medium and natural data case (Appendix \ref{app:natural}), this study's findings primarily rely on the Pythia model suite and synthetic/fine-tuning datasets, which are chosen to ensure architectural consistency and data purity for controlled analysis. It remains to be validated whether these conclusions can be directly generalized to other model architectures and more complex, noisy real-world data.

    \item The current work concentrates on direct knowledge interference arising from imbalanced knowledge popularity. The applicability of \framework to practical tasks like multi-hop reasoning and code generation, remains a key area for future exploration.

    \item Performing dynamic analysis of knowledge circuits throughout the entire training process is computationally intensive, which could potentially hinder its scalability to very large models or extensive training runs. Future work aims to develop more computationally efficient techniques, such as more efficient circuit extraction methods like Information Flow Routes (IFR), to mitigate this challenge.

    % \item The proposed circuit-based recovery is demonstrated for specific instances, and generalizing it into a robust, model-wide mitigation strategy remains challenging. We will explore methods to translate instance-specific circuit optimizations into broader strategies, potentially through novel fine-tuning or regularization techniques.
\end{enumerate}

\section*{Acknowledgments}

This work was supported by the National Natural Science Foundation of China (Grant No.62506318); Guangdong Provincial Department of Education Project (Grant No.2024KQNCX028); CAAI-Ant Group Research Fund; Scientific Research Projects for the Higher-educational Institutions (Grant No.2024312096), Education Bureau of Guangzhou Municipality; Guangzhou-HKUST(GZ) Joint Funding Program (Grant No.2025A03J3957), Education Bureau of Guangzhou Municipality.

% \clearpage
\bibliography{custom}

\begin{thebibliography}{70}
\providecommand{\natexlab}[1]{#1}

\bibitem[{Asgari et~al.(2025)Asgari, Monta{\~n}a-Brown, Dubois, Khalil, Balloch, Yeung, and Pimenta}]{asgari2025framework}
Elham Asgari, Nina Monta{\~n}a-Brown, Magda Dubois, Saleh Khalil, Jasmine Balloch, Joshua~Au Yeung, and Dominic Pimenta. 2025.
\newblock A framework to assess clinical safety and hallucination rates of llms for medical text summarisation.
\newblock \emph{npj Digital Medicine}, 8(1):1--15.

\bibitem[{Biderman et~al.(2023)Biderman, Schoelkopf, Anthony, Bradley, O’Brien, Hallahan, Khan, Purohit, Prashanth, Raff et~al.}]{biderman2023pythia}
Stella Biderman, Hailey Schoelkopf, Quentin~Gregory Anthony, Herbie Bradley, Kyle O’Brien, Eric Hallahan, Mohammad~Aflah Khan, Shivanshu Purohit, USVSN~Sai Prashanth, Edward Raff, and 1 others. 2023.
\newblock Pythia: A suite for analyzing large language models across training and scaling.
\newblock In \emph{International Conference on Machine Learning}, pages 2397--2430. PMLR.

\bibitem[{Brown et~al.(2020)Brown, Mann, Ryder, Subbiah, Kaplan, Dhariwal, Neelakantan, Shyam, Sastry, Askell et~al.}]{brown2020language}
Tom Brown, Benjamin Mann, Nick Ryder, Melanie Subbiah, Jared~D Kaplan, Prafulla Dhariwal, Arvind Neelakantan, Pranav Shyam, Girish Sastry, Amanda Askell, and 1 others. 2020.
\newblock Language models are few-shot learners.
\newblock \emph{Advances in neural information processing systems}, 33:1877--1901.

\bibitem[{Cai et~al.(2024)Cai, Cao, Chen, Chen, Chen, Chen, Chen, Chen, Chen, Chu, Dong, Duan, Fan, Fei, Gao, Ge, Gu, Gu, Gui, Guo, Guo, He, Hu, Huang, Jiang, Jiao, Jin, Lei, Li, Li, Li, Li, Li, Li, Liu, Liu, Hong, Liu, Liu, Liu, Lv, Lv, Lv, Ma, Ma, Ma, Ning, Ouyang, Qiu, Qu, Shang, Shao, Song, Song, Sui, Sun, Sun, Tang, Wang, Wang, Wang, Wang, Wang, Wang, Wang, Wei, Weng, Wu, Xiong, Xu, Xu, Yan, Yan, Yang, Ye, Ying, Yu, Yu, Zang, Zhang, Zhang, Zhang, Zhang, Zhang, Zhang, Zhang, Zhang, Zhang, Zhang, Zhang, Zhao, Zhao, Zhao, Zhou, Zhou, Zhuo, Zou, Qiu, Qiao, and Lin}]{cai2024internlm2}
Zheng Cai, Maosong Cao, Haojiong Chen, Kai Chen, Keyu Chen, Xin Chen, Xun Chen, Zehui Chen, Zhi Chen, Pei Chu, Xiaoyi Dong, Haodong Duan, Qi~Fan, Zhaoye Fei, Yang Gao, Jiaye Ge, Chenya Gu, Yuzhe Gu, Tao Gui, and 81 others. 2024.
\newblock \href {https://arxiv.org/abs/2403.17297} {Internlm2 technical report}.
\newblock \emph{Preprint}, arXiv:2403.17297.

\bibitem[{Chakraborty et~al.(2025)Chakraborty, Ornik, and Driggs-Campbell}]{chakraborty2025hallucination}
Neeloy Chakraborty, Melkior Ornik, and Katherine Driggs-Campbell. 2025.
\newblock Hallucination detection in foundation models for decision-making: A flexible definition and review of the state of the art.
\newblock \emph{ACM Computing Surveys}.

\bibitem[{Chang et~al.(2024)Chang, Wang, Wang, Wu, Yang, Zhu, Chen, Yi, Wang, Wang et~al.}]{chang2024survey}
Yupeng Chang, Xu~Wang, Jindong Wang, Yuan Wu, Linyi Yang, Kaijie Zhu, Hao Chen, Xiaoyuan Yi, Cunxiang Wang, Yidong Wang, and 1 others. 2024.
\newblock A survey on evaluation of large language models.
\newblock \emph{ACM transactions on intelligent systems and technology}, 15(3):1--45.

\bibitem[{Chern et~al.(2023)Chern, Chern, Chen, Yuan, Feng, Zhou, He, Neubig, Liu et~al.}]{chern2023factool}
I~Chern, Steffi Chern, Shiqi Chen, Weizhe Yuan, Kehua Feng, Chunting Zhou, Junxian He, Graham Neubig, Pengfei Liu, and 1 others. 2023.
\newblock Factool: Factuality detection in generative ai--a tool augmented framework for multi-task and multi-domain scenarios.
\newblock \emph{arXiv preprint arXiv:2307.13528}.

\bibitem[{Chu et~al.(2025)Chu, Wang, Xie, Zhu, Yan, Ye, Zhong, Hu, Liang, Yu et~al.}]{chu2025llm}
Zhendong Chu, Shen Wang, Jian Xie, Tinghui Zhu, Yibo Yan, Jinheng Ye, Aoxiao Zhong, Xuming Hu, Jing Liang, Philip~S Yu, and 1 others. 2025.
\newblock Llm agents for education: Advances and applications.
\newblock \emph{arXiv preprint arXiv:2503.11733}.

\bibitem[{Conmy et~al.(2023)Conmy, Mavor-Parker, Lynch, Heimersheim, and Garriga-Alonso}]{conmy2023towards}
Arthur Conmy, Augustine Mavor-Parker, Aengus Lynch, Stefan Heimersheim, and Adri{\`a} Garriga-Alonso. 2023.
\newblock Towards automated circuit discovery for mechanistic interpretability.
\newblock \emph{Advances in Neural Information Processing Systems}, 36:16318--16352.

\bibitem[{Dang et~al.(2024)Dang, Gao, Yan, Zou, Gu, Liu, and Hu}]{dang2024exploring}
Yunkai Dang, Mengxi Gao, Yibo Yan, Xin Zou, Yanggan Gu, Aiwei Liu, and Xuming Hu. 2024.
\newblock Exploring response uncertainty in mllms: An empirical evaluation under misleading scenarios.
\newblock \emph{arXiv preprint arXiv:2411.02708}.

\bibitem[{Dhuliawala et~al.(2023)Dhuliawala, Komeili, Xu, Raileanu, Li, Celikyilmaz, and Weston}]{dhuliawala2023chain}
Shehzaad Dhuliawala, Mojtaba Komeili, Jing Xu, Roberta Raileanu, Xian Li, Asli Celikyilmaz, and Jason Weston. 2023.
\newblock Chain-of-verification reduces hallucination in large language models.
\newblock \emph{arXiv preprint arXiv:2309.11495}.

\bibitem[{Dong et~al.(2025)Dong, Zhu, Zhang, Wang, Wen, and Dou}]{dong2025understand}
Guanting Dong, Yutao Zhu, Chenghao Zhang, Zechen Wang, Ji-Rong Wen, and Zhicheng Dou. 2025.
\newblock Understand what llm needs: Dual preference alignment for retrieval-augmented generation.
\newblock In \emph{Proceedings of the ACM on Web Conference 2025}, pages 4206--4225.

\bibitem[{Dong et~al.(2022)Dong, Li, Dai, Zheng, Ma, Li, Xia, Xu, Wu, Liu et~al.}]{dong2022survey}
Qingxiu Dong, Lei Li, Damai Dai, Ce~Zheng, Jingyuan Ma, Rui Li, Heming Xia, Jingjing Xu, Zhiyong Wu, Tianyu Liu, and 1 others. 2022.
\newblock A survey on in-context learning.
\newblock \emph{arXiv preprint arXiv:2301.00234}.

\bibitem[{Elhage et~al.(2021)Elhage, Nanda, Olsson, Henighan, Joseph, Mann, Askell, Bai, Chen, Conerly et~al.}]{elhage2021mathematical}
Nelson Elhage, Neel Nanda, Catherine Olsson, Tom Henighan, Nicholas Joseph, Ben Mann, Amanda Askell, Yuntao Bai, Anna Chen, Tom Conerly, and 1 others. 2021.
\newblock A mathematical framework for transformer circuits.
\newblock \emph{Transformer Circuits Thread}, 1(1):12.

\bibitem[{Grattafiori et~al.(2024)Grattafiori, Dubey, Jauhri, Pandey, Kadian, Al-Dahle, Letman, Mathur, Schelten, Vaughan et~al.}]{grattafiori2024llama}
Aaron Grattafiori, Abhimanyu Dubey, Abhinav Jauhri, Abhinav Pandey, Abhishek Kadian, Ahmad Al-Dahle, Aiesha Letman, Akhil Mathur, Alan Schelten, Alex Vaughan, and 1 others. 2024.
\newblock The llama 3 herd of models.
\newblock \emph{arXiv preprint arXiv:2407.21783}.

\bibitem[{Guo et~al.(2025)Guo, Yang, Zhang, Song, Zhang, Xu, Zhu, Ma, Wang, Bi et~al.}]{guo2025deepseek}
Daya Guo, Dejian Yang, Haowei Zhang, Junxiao Song, Ruoyu Zhang, Runxin Xu, Qihao Zhu, Shirong Ma, Peiyi Wang, Xiao Bi, and 1 others. 2025.
\newblock Deepseek-r1: Incentivizing reasoning capability in llms via reinforcement learning.
\newblock \emph{arXiv preprint arXiv:2501.12948}.

\bibitem[{Hakimi et~al.(2025)Hakimi, Modarressi, Wicke, and Schütze}]{hakimi2025timecoursemechinterpanalyzing}
Ahmad~Dawar Hakimi, Ali Modarressi, Philipp Wicke, and Hinrich Schütze. 2025.
\newblock \href {https://arxiv.org/abs/2506.03434} {Time course mechinterp: Analyzing the evolution of components and knowledge in large language models}.
\newblock \emph{Preprint}, arXiv:2506.03434.

\bibitem[{Hanna et~al.(2024)Hanna, Pezzelle, and Belinkov}]{hanna2024have}
Michael Hanna, Sandro Pezzelle, and Yonatan Belinkov. 2024.
\newblock Have faith in faithfulness: Going beyond circuit overlap when finding model mechanisms.
\newblock \emph{arXiv preprint arXiv:2403.17806}.

\bibitem[{Huang et~al.(2024{\natexlab{a}})Huang, Huo, Yan, Wang, Yue, and Hu}]{huang2024miner}
Kaichen Huang, Jiahao Huo, Yibo Yan, Kun Wang, Yutao Yue, and Xuming Hu. 2024{\natexlab{a}}.
\newblock Miner: Mining the underlying pattern of modality-specific neurons in multimodal large language models.
\newblock \emph{arXiv preprint arXiv:2410.04819}.

\bibitem[{Huang et~al.(2025)Huang, Yu, Ma, Zhong, Feng, Wang, Chen, Peng, Feng, Qin et~al.}]{huang2025survey}
Lei Huang, Weijiang Yu, Weitao Ma, Weihong Zhong, Zhangyin Feng, Haotian Wang, Qianglong Chen, Weihua Peng, Xiaocheng Feng, Bing Qin, and 1 others. 2025.
\newblock A survey on hallucination in large language models: Principles, taxonomy, challenges, and open questions.
\newblock \emph{ACM Transactions on Information Systems}, 43(2):1--55.

\bibitem[{Huang et~al.(2024{\natexlab{b}})Huang, Panwar, Goyal, and Hahn}]{huang2024inversionview}
Xinting Huang, Madhur Panwar, Navin Goyal, and Michael Hahn. 2024{\natexlab{b}}.
\newblock Inversionview: A general-purpose method for reading information from neural activations.
\newblock \emph{arXiv preprint arXiv:2405.17653}.

\bibitem[{Huo et~al.(2024)Huo, Yan, Hu, Yue, and Hu}]{huo2024mmneuron}
Jiahao Huo, Yibo Yan, Boren Hu, Yutao Yue, and Xuming Hu. 2024.
\newblock Mmneuron: Discovering neuron-level domain-specific interpretation in multimodal large language model.
\newblock \emph{arXiv preprint arXiv:2406.11193}.

\bibitem[{Jaech et~al.(2024)Jaech, Kalai, Lerer, Richardson, El-Kishky, Low, Helyar, Madry, Beutel, Carney et~al.}]{jaech2024openaio1}
Aaron Jaech, Adam Kalai, Adam Lerer, Adam Richardson, Ahmed El-Kishky, Aiden Low, Alec Helyar, Aleksander Madry, Alex Beutel, Alex Carney, and 1 others. 2024.
\newblock Openai o1 system card.
\newblock \emph{arXiv preprint arXiv:2412.16720}.

\bibitem[{Kadavath et~al.(2022)Kadavath, Conerly, Askell, Henighan, Drain, Perez, Schiefer, Hatfield-Dodds, DasSarma, Tran-Johnson et~al.}]{kadavath2022language}
Saurav Kadavath, Tom Conerly, Amanda Askell, Tom Henighan, Dawn Drain, Ethan Perez, Nicholas Schiefer, Zac Hatfield-Dodds, Nova DasSarma, Eli Tran-Johnson, and 1 others. 2022.
\newblock Language models (mostly) know what they know.
\newblock \emph{arXiv preprint arXiv:2207.05221}.

\bibitem[{Kim et~al.(2024)Kim, Lee, and Mutlu}]{kim2024understanding}
Callie~Y Kim, Christine~P Lee, and Bilge Mutlu. 2024.
\newblock Understanding large-language model (llm)-powered human-robot interaction.
\newblock In \emph{Proceedings of the 2024 ACM/IEEE international conference on human-robot interaction}, pages 371--380.

\bibitem[{Li et~al.(2025)Li, Zhang, Zhang, Zhang, Liu, Yao, Xu, Zheng, Wang, Chen et~al.}]{li2025system1}
Zhong-Zhi Li, Duzhen Zhang, Ming-Liang Zhang, Jiaxin Zhang, Zengyan Liu, Yuxuan Yao, Haotian Xu, Junhao Zheng, Pei-Jie Wang, Xiuyi Chen, and 1 others. 2025.
\newblock From system 1 to system 2: A survey of reasoning large language models.
\newblock \emph{arXiv preprint arXiv:2502.17419}.

\bibitem[{Luo et~al.(2023)Luo, Xiao, and Ma}]{luo2023zero}
Junyu Luo, Cao Xiao, and Fenglong Ma. 2023.
\newblock Zero-resource hallucination prevention for large language models.
\newblock \emph{arXiv preprint arXiv:2309.02654}.

\bibitem[{Marks et~al.(2024)Marks, Rager, Michaud, Belinkov, Bau, and Mueller}]{marks2024sparse}
Samuel Marks, Can Rager, Eric~J Michaud, Yonatan Belinkov, David Bau, and Aaron Mueller. 2024.
\newblock Sparse feature circuits: Discovering and editing interpretable causal graphs in language models.
\newblock \emph{arXiv preprint arXiv:2403.19647}.

\bibitem[{Masterman et~al.(2024)Masterman, Besen, Sawtell, and Chao}]{masterman2024landscape}
Tula Masterman, Sandi Besen, Mason Sawtell, and Alex Chao. 2024.
\newblock The landscape of emerging ai agent architectures for reasoning, planning, and tool calling: A survey.
\newblock \emph{arXiv preprint arXiv:2404.11584}.

\bibitem[{Min et~al.(2023)Min, Krishna, Lyu, Lewis, Yih, Koh, Iyyer, Zettlemoyer, and Hajishirzi}]{min2023factscore}
Sewon Min, Kalpesh Krishna, Xinxi Lyu, Mike Lewis, Wen-tau Yih, Pang~Wei Koh, Mohit Iyyer, Luke Zettlemoyer, and Hannaneh Hajishirzi. 2023.
\newblock Factscore: Fine-grained atomic evaluation of factual precision in long form text generation.
\newblock \emph{arXiv preprint arXiv:2305.14251}.

\bibitem[{Nam et~al.(2024)Nam, Macvean, Hellendoorn, Vasilescu, and Myers}]{nam2024using}
Daye Nam, Andrew Macvean, Vincent Hellendoorn, Bogdan Vasilescu, and Brad Myers. 2024.
\newblock Using an llm to help with code understanding.
\newblock In \emph{Proceedings of the IEEE/ACM 46th International Conference on Software Engineering}, pages 1--13.

\bibitem[{nostalgebraist(2020)}]{nostalgebraist2020interpreting}
nostalgebraist. 2020.
\newblock Interpreting {GPT}: the logit lens.
\newblock \url{https://www.lesswrong.com/posts/AcKRB8wDpdaN6v6ru/interpreting-gpt-the-logit-lens}.

\bibitem[{Olah et~al.(2020)Olah, Cammarata, Schubert, Goh, Petrov, and Carter}]{olah2020zoom}
Chris Olah, Nick Cammarata, Ludwig Schubert, Gabriel Goh, Michael Petrov, and Shan Carter. 2020.
\newblock Zoom in: An introduction to circuits.
\newblock \emph{Distill}, 5(3):e00024--001.

\bibitem[{Ou et~al.(2025)Ou, Yao, Zhang, Jin, Sun, Deng, Li, and Chen}]{ou2025llms}
Yixin Ou, Yunzhi Yao, Ningyu Zhang, Hui Jin, Jiacheng Sun, Shumin Deng, Zhenguo Li, and Huajun Chen. 2025.
\newblock How do llms acquire new knowledge? a knowledge circuits perspective on continual pre-training.
\newblock \emph{arXiv preprint arXiv:2502.11196}.

\bibitem[{Pan et~al.(2025)Pan, Wang, and Li}]{pan2025understanding}
Zhixuan Pan, Shaowen Wang, and Jian Li. 2025.
\newblock Understanding llm behaviors via compression: Data generation, knowledge acquisition and scaling laws.
\newblock \emph{arXiv preprint arXiv:2504.09597}.

\bibitem[{Putta et~al.(2024)Putta, Mills, Garg, Motwani, Finn, Garg, and Rafailov}]{putta2024agent}
Pranav Putta, Edmund Mills, Naman Garg, Sumeet Motwani, Chelsea Finn, Divyansh Garg, and Rafael Rafailov. 2024.
\newblock Agent q: Advanced reasoning and learning for autonomous ai agents.
\newblock \emph{arXiv preprint arXiv:2408.07199}.

\bibitem[{Rai et~al.(2024)Rai, Zhou, Feng, Saparov, and Yao}]{rai2024practical}
Daking Rai, Yilun Zhou, Shi Feng, Abulhair Saparov, and Ziyu Yao. 2024.
\newblock A practical review of mechanistic interpretability for transformer-based language models.
\newblock \emph{arXiv preprint arXiv:2407.02646}.

\bibitem[{Rawte et~al.(2023)Rawte, Sheth, and Das}]{rawte2023survey}
Vipula Rawte, Amit Sheth, and Amitava Das. 2023.
\newblock A survey of hallucination in large foundation models.
\newblock \emph{arXiv preprint arXiv:2309.05922}.

\bibitem[{Su et~al.(2025)Su, Yan, Fu, Zhang, Ye, Liu, Huo, Zhou, and Hu}]{su2025essayjudge}
Jiamin Su, Yibo Yan, Fangteng Fu, Han Zhang, Jingheng Ye, Xiang Liu, Jiahao Huo, Huiyu Zhou, and Xuming Hu. 2025.
\newblock Essayjudge: A multi-granular benchmark for assessing automated essay scoring capabilities of multimodal large language models.
\newblock \emph{arXiv preprint arXiv:2502.11916}.

\bibitem[{Suzgun et~al.(2022)Suzgun, Scales, Sch{\"a}rli, Gehrmann, Tay, Chung, Chowdhery, Le, Chi, Zhou, , and Wei}]{suzgun2022challenging}
Mirac Suzgun, Nathan Scales, Nathanael Sch{\"a}rli, Sebastian Gehrmann, Yi~Tay, Hyung~Won Chung, Aakanksha Chowdhery, Quoc~V Le, Ed~H Chi, Denny Zhou, , and Jason Wei. 2022.
\newblock Challenging big-bench tasks and whether chain-of-thought can solve them.
\newblock \emph{arXiv preprint arXiv:2210.09261}.

\bibitem[{Syed et~al.(2023)Syed, Rager, and Conmy}]{syed2023attribution}
Aaquib Syed, Can Rager, and Arthur Conmy. 2023.
\newblock Attribution patching outperforms automated circuit discovery.
\newblock \emph{arXiv preprint arXiv:2310.10348}.

\bibitem[{Tang et~al.(2024)Tang, Luo, Huang, Zhang, Wang, Zhao, Wei, and Wen}]{tang2024language}
Tianyi Tang, Wenyang Luo, Haoyang Huang, Dongdong Zhang, Xiaolei Wang, Xin Zhao, Furu Wei, and Ji-Rong Wen. 2024.
\newblock Language-specific neurons: The key to multilingual capabilities in large language models.
\newblock \emph{arXiv preprint arXiv:2402.16438}.

\bibitem[{Team(2024)}]{qwen2.5}
Qwen Team. 2024.
\newblock \href {https://qwenlm.github.io/blog/qwen2.5/} {Qwen2.5: A party of foundation models}.

\bibitem[{Trung et~al.(2024)Trung, Zhang, Jie, Sun, Jin, and Li}]{trung2024reft}
Luong Trung, Xinbo Zhang, Zhanming Jie, Peng Sun, Xiaoran Jin, and Hang Li. 2024.
\newblock Reft: Reasoning with reinforced fine-tuning.
\newblock In \emph{Proceedings of the 62nd Annual Meeting of the Association for Computational Linguistics (Volume 1: Long Papers)}, pages 7601--7614.

\bibitem[{Varshney et~al.(2023)Varshney, Yao, Zhang, Chen, and Yu}]{varshney2023stitch}
Neeraj Varshney, Wenlin Yao, Hongming Zhang, Jianshu Chen, and Dong Yu. 2023.
\newblock A stitch in time saves nine: Detecting and mitigating hallucinations of llms by validating low-confidence generation.
\newblock \emph{arXiv preprint arXiv:2307.03987}.

\bibitem[{Wang et~al.(2022)Wang, Variengien, Conmy, Shlegeris, and Steinhardt}]{wang2022interpretability}
Kevin Wang, Alexandre Variengien, Arthur Conmy, Buck Shlegeris, and Jacob Steinhardt. 2022.
\newblock Interpretability in the wild: a circuit for indirect object identification in gpt-2 small.
\newblock \emph{arXiv preprint arXiv:2211.00593}.

\bibitem[{Wang et~al.(2025)Wang, Zhang, Zhou, Wu, Yu, Zhao, Yin, Fu, Yan, Luo et~al.}]{wang2025comprehensive}
Kun Wang, Guibin Zhang, Zhenhong Zhou, Jiahao Wu, Miao Yu, Shiqian Zhao, Chenlong Yin, Jinhu Fu, Yibo Yan, Hanjun Luo, and 1 others. 2025.
\newblock A comprehensive survey in llm (-agent) full stack safety: Data, training and deployment.
\newblock \emph{arXiv preprint arXiv:2504.15585}.

\bibitem[{Wang et~al.(2023)Wang, Li, Dai, Chen, Zhou, Meng, Zhou, and Sun}]{wang2023label}
Lean Wang, Lei Li, Damai Dai, Deli Chen, Hao Zhou, Fandong Meng, Jie Zhou, and Xu~Sun. 2023.
\newblock Label words are anchors: An information flow perspective for understanding in-context learning.
\newblock \emph{arXiv preprint arXiv:2305.14160}.

\bibitem[{Webb et~al.(2023)Webb, Holyoak, and Lu}]{webb2023emergent}
Taylor Webb, Keith~J Holyoak, and Hongjing Lu. 2023.
\newblock Emergent analogical reasoning in large language models.
\newblock \emph{Nature Human Behaviour}, 7(9):1526--1541.

\bibitem[{Wendler et~al.(2024)Wendler, Veselovsky, Monea, and West}]{wendler2024llamas}
Chris Wendler, Veniamin Veselovsky, Giovanni Monea, and Robert West. 2024.
\newblock Do llamas work in english? on the latent language of multilingual transformers.
\newblock In \emph{Proceedings of the 62nd Annual Meeting of the Association for Computational Linguistics (Volume 1: Long Papers)}, pages 15366--15394.

\bibitem[{Xun et~al.(2025)Xun, Tao, Li, Shi, Lin, Zhu, Yan, Li, Zhang, Wang et~al.}]{xun2025rtv}
Shuhang Xun, Sicheng Tao, Jungang Li, Yibo Shi, Zhixin Lin, Zhanhui Zhu, Yibo Yan, Hanqian Li, Linghao Zhang, Shikang Wang, and 1 others. 2025.
\newblock Rtv-bench: Benchmarking mllm continuous perception, understanding and reasoning through real-time video.
\newblock \emph{arXiv preprint arXiv:2505.02064}.

\bibitem[{Yan and Lee(2024)}]{yan2024georeasoner}
Yibo Yan and Joey Lee. 2024.
\newblock Georeasoner: Reasoning on geospatially grounded context for natural language understanding.
\newblock In \emph{Proceedings of the 33rd ACM International Conference on Information and Knowledge Management}, pages 4163--4167.

\bibitem[{Yan et~al.(2024{\natexlab{a}})Yan, Su, He, Fu, Zheng, Lyu, Wang, Wang, Wen, and Hu}]{yan2024survey}
Yibo Yan, Jiamin Su, Jianxiang He, Fangteng Fu, Xu~Zheng, Yuanhuiyi Lyu, Kun Wang, Shen Wang, Qingsong Wen, and Xuming Hu. 2024{\natexlab{a}}.
\newblock A survey of mathematical reasoning in the era of multimodal large language model: Benchmark, method \& challenges.
\newblock \emph{arXiv preprint arXiv:2412.11936}.

\bibitem[{Yan et~al.(2024{\natexlab{b}})Yan, Wang, Huo, Li, Li, Su, Gao, Zhang, Xu, Chu et~al.}]{yan2024errorradar}
Yibo Yan, Shen Wang, Jiahao Huo, Hang Li, Boyan Li, Jiamin Su, Xiong Gao, Yi-Fan Zhang, Tianlong Xu, Zhendong Chu, and 1 others. 2024{\natexlab{b}}.
\newblock Errorradar: Benchmarking complex mathematical reasoning of multimodal large language models via error detection.
\newblock \emph{arXiv preprint arXiv:2410.04509}.

\bibitem[{Yan et~al.(2025{\natexlab{a}})Yan, Wang, Huo, Ye, Chu, Hu, Yu, Gomes, Selman, and Wen}]{yan2025position}
Yibo Yan, Shen Wang, Jiahao Huo, Jingheng Ye, Zhendong Chu, Xuming Hu, Philip~S Yu, Carla Gomes, Bart Selman, and Qingsong Wen. 2025{\natexlab{a}}.
\newblock Position: Multimodal large language models can significantly advance scientific reasoning.
\newblock \emph{arXiv preprint arXiv:2502.02871}.

\bibitem[{Yan et~al.(2025{\natexlab{b}})Yan, Wang, Huo, Yu, Hu, and Wen}]{yan2025mathagent}
Yibo Yan, Shen Wang, Jiahao Huo, Philip~S Yu, Xuming Hu, and Qingsong Wen. 2025{\natexlab{b}}.
\newblock Mathagent: Leveraging a mixture-of-math-agent framework for real-world multimodal mathematical error detection.
\newblock \emph{arXiv preprint arXiv:2503.18132}.

\bibitem[{Yan et~al.(2024{\natexlab{c}})Yan, Wen, Zhong, Chen, Chen, Wen, Zimmermann, and Liang}]{yan2024urbanclip}
Yibo Yan, Haomin Wen, Siru Zhong, Wei Chen, Haodong Chen, Qingsong Wen, Roger Zimmermann, and Yuxuan Liang. 2024{\natexlab{c}}.
\newblock Urbanclip: Learning text-enhanced urban region profiling with contrastive language-image pretraining from the web.
\newblock In \emph{Proceedings of the ACM Web Conference 2024}, pages 4006--4017.

\bibitem[{Yao et~al.(2023)Yao, Ning, Liu, Ning, Liu, and Yuan}]{yao2023llm}
Jia-Yu Yao, Kun-Peng Ning, Zhen-Hui Liu, Mu-Nan Ning, Yu-Yang Liu, and Li~Yuan. 2023.
\newblock Llm lies: Hallucinations are not bugs, but features as adversarial examples.
\newblock \emph{arXiv preprint arXiv:2310.01469}.

\bibitem[{Yao et~al.(2024)Yao, Zhang, Xi, Wang, Xu, Deng, and Chen}]{yao2024knowledge}
Yunzhi Yao, Ningyu Zhang, Zekun Xi, Mengru Wang, Ziwen Xu, Shumin Deng, and Huajun Chen. 2024.
\newblock Knowledge circuits in pretrained transformers.
\newblock \emph{arXiv preprint arXiv:2405.17969}.

\bibitem[{Yue(2025)}]{yue2025survey}
Murong Yue. 2025.
\newblock A survey of large language model agents for question answering.
\newblock \emph{arXiv preprint arXiv:2503.19213}.

\bibitem[{Zhang et~al.(2024{\natexlab{a}})Zhang, Peng, Tian, Zhou, Jin, Song, Mi, and Meng}]{zhang2024self}
Xiaoying Zhang, Baolin Peng, Ye~Tian, Jingyan Zhou, Lifeng Jin, Linfeng Song, Haitao Mi, and Helen Meng. 2024{\natexlab{a}}.
\newblock Self-alignment for factuality: Mitigating hallucinations in llms via self-evaluation.
\newblock \emph{arXiv preprint arXiv:2402.09267}.

\bibitem[{Zhang et~al.(2024{\natexlab{b}})Zhang, Li, Liu, Yu, Fung, Li, Li, and Ji}]{zhang2024knowledge}
Yuji Zhang, Sha Li, Jiateng Liu, Pengfei Yu, Yi~R Fung, Jing Li, Manling Li, and Heng Ji. 2024{\natexlab{b}}.
\newblock Knowledge overshadowing causes amalgamated hallucination in large language models.
\newblock \emph{arXiv preprint arXiv:2407.08039}.

\bibitem[{Zhang et~al.(2025{\natexlab{a}})Zhang, Li, Qian, Liu, Yu, Han, Fung, McKeown, Zhai, Li et~al.}]{zhang2025law}
Yuji Zhang, Sha Li, Cheng Qian, Jiateng Liu, Pengfei Yu, Chi Han, Yi~R Fung, Kathleen McKeown, Chengxiang Zhai, Manling Li, and 1 others. 2025{\natexlab{a}}.
\newblock The law of knowledge overshadowing: Towards understanding, predicting, and preventing llm hallucination.
\newblock \emph{arXiv preprint arXiv:2502.16143}.

\bibitem[{Zhang et~al.(2025{\natexlab{b}})Zhang, Yao, Zhang, Tang, Ma, He, Wang, Gerstein, Wang, Liu et~al.}]{zhang2025igniting}
Zhuosheng Zhang, Yao Yao, Aston Zhang, Xiangru Tang, Xinbei Ma, Zhiwei He, Yiming Wang, Mark Gerstein, Rui Wang, Gongshen Liu, and 1 others. 2025{\natexlab{b}}.
\newblock Igniting language intelligence: The hitchhiker’s guide from chain-of-thought reasoning to language agents.
\newblock \emph{ACM Computing Surveys}, 57(8):1--39.

\bibitem[{Zhao et~al.(2024)Zhao, Chen, Yang, Liu, Deng, Cai, Wang, Yin, and Du}]{zhao2024explainability}
Haiyan Zhao, Hanjie Chen, Fan Yang, Ninghao Liu, Huiqi Deng, Hengyi Cai, Shuaiqiang Wang, Dawei Yin, and Mengnan Du. 2024.
\newblock Explainability for large language models: A survey.
\newblock \emph{ACM Transactions on Intelligent Systems and Technology}, 15(2):1--38.

\bibitem[{Zheng et~al.(2024)Zheng, Chen, Yan, Zou, and Hu}]{zheng2024reefknot}
Kening Zheng, Junkai Chen, Yibo Yan, Xin Zou, and Xuming Hu. 2024.
\newblock Reefknot: A comprehensive benchmark for relation hallucination evaluation, analysis and mitigation in multimodal large language models.
\newblock \emph{arXiv preprint arXiv:2408.09429}.

\bibitem[{Zhou et~al.(2024)Zhou, Yan, Zou, Wang, Liu, and Hu}]{zhou2024mitigating}
Guanyu Zhou, Yibo Yan, Xin Zou, Kun Wang, Aiwei Liu, and Xuming Hu. 2024.
\newblock Mitigating modality prior-induced hallucinations in multimodal large language models via deciphering attention causality.
\newblock \emph{arXiv preprint arXiv:2410.04780}.

\bibitem[{Zhu et~al.(2024)Zhu, Liu, Yu, Tang, Yan, Li, Xiong, Xu, and Blaschko}]{zhu2024fastmem}
Junyi Zhu, Shuochen Liu, Yu~Yu, Bo~Tang, Yibo Yan, Zhiyu Li, Feiyu Xiong, Tong Xu, and Matthew~B Blaschko. 2024.
\newblock Fastmem: fast memorization of prompt improves context awareness of large language models.
\newblock \emph{arXiv preprint arXiv:2406.16069}.

\bibitem[{Zou et~al.(2024)Zou, Wang, Yan, Huang, Zheng, Chen, Tang, and Hu}]{zou2024look}
Xin Zou, Yizhou Wang, Yibo Yan, Sirui Huang, Kening Zheng, Junkai Chen, Chang Tang, and Xuming Hu. 2024.
\newblock Look twice before you answer: Memory-space visual retracing for hallucination mitigation in multimodal large language models.
\newblock \emph{arXiv preprint arXiv:2410.03577}.

\bibitem[{Zucchet et~al.(2025)Zucchet, Bornschein, Chan, Lampinen, Pascanu, and De}]{zucchet2025languagemodelslearnfacts}
Nicolas Zucchet, Jörg Bornschein, Stephanie Chan, Andrew Lampinen, Razvan Pascanu, and Soham De. 2025.
\newblock \href {https://arxiv.org/abs/2503.21676} {How do language models learn facts? dynamics, curricula and hallucinations}.
\newblock \emph{Preprint}, arXiv:2503.21676.

\end{thebibliography}
\clearpage
\appendix

\section{More Related Work}
\label{app:more_related_work}

\subsection{Large Language Models}
\label{app:llm}
First proposed by~\citet{brown2020language}, Transformer-based auto-regressive LLMs have demonstrated strong performance across a variety of NLP tasks, including question answering~\cite{yue2025survey}, in-context learning~\cite{dong2022survey}, and analogical reasoning~\cite{webb2023emergent}. pre-trained on large-scale text corpora, LLMs have acquired extensive real-world knowledge from web sources. As a result, models such as InternLM2.5~\cite{cai2024internlm2}, Qwen2.5~\cite{qwen2.5}, and LLaMA3.3~\cite{grattafiori2024llama} have shown excellent performance on world knowledge benchmarks~\cite{suzgun2022challenging}. Therefore, over the past year, LLMs have demonstrated remarkable capabilities in understanding-related tasks across various fields \cite{kim2024understanding,nam2024using,yan2024urbanclip,yan2024georeasoner,yan2024errorradar,dong2025understand,su2025essayjudge}.\par
Recently, there has been a growing trend toward enhancing LLMs’ reasoning capabilities on complex tasks~\cite{guo2025deepseek,jaech2024openaio1} by generating long Chain-of-Thoughts (CoTs), with reinforcement learning (RL) emerging as an effective tool to encourage this behavior~\cite{li2025system1,trung2024reft}. Recently, there have also been efforts to explore collaboration between LLMs to enhance their reasoning abilities \cite{zhang2025igniting,putta2024agent,masterman2024landscape,yan2025mathagent,chu2025llm}.\par
Despite these advancements, existing LLMs still suffer from factual hallucinations in practice \cite{pan2025understanding,asgari2025framework}, with knowledge overshadowing identified as a primary contributing factor~\cite{zhang2024knowledge}. While existing interoperability works make great efforts on the mechanism of LLM training and generating~\cite{zhao2024explainability}, most of them solely focus on isolated model versions like GPT2~\cite{wang2023label} and LLaMA2~\cite{wendler2024llamas, tang2024language}.\par
In this paper, we utilize the Pythia suite~\cite{biderman2023pythia} to investigate the evolution and underlying mechanisms of knowledge overshadowing across models of varying sizes: 70M, 410M, 1.4B, and 2.8B parameters. Sharing a unified architecture, this model suite eliminates design variability, thereby providing clearer and more reliable insights into the scaling behavior of the knowledge overshadowing phenomenon in LLMs.

\section{\framework Details}

%%%%%%%%%%%%%%%%%具体的知识电路构建%%%%%%%%%%%%%%%%%%%
\subsection{Circuit Construction}
\label{app:construction}
Knowledge circuit is as a sparse computational subgraph within the LLMs. The construction of such a circuit involves identifying and retaining the most influential components (nodes, including MLPs and attention heads) and connections (edges) while pruning less critical ones.

We adapted the optimized circuit construction method provided by \cite{yao2024knowledge}. The process begins by representing the LLM as a directed acyclic graph (DAG), $G=(V, E)$, where $V$ encompasses input embeddings, attention heads, MLP layers, and output logits, and $E$ represents the information flow between these components. Our goal is to identify a subgraph $C$ that is critical for recognizing the key component of a given input prompt. In the context of knowledge overshadowing, this component is $\{X_{dom}, X_{sub}\}$, which represents the difference between $P_{dom}$ and $P_{sub}$.

The adapted construction method is similar to Edge Attribution Patching  with Integrated Gradients (EAP-IG) \cite{hanna2024have}, which involves:

\begin{enumerate}
    \item  \textbf{Paired Inputs:} For a given background $X_{bg}$, we create two primary input prompts: $P_{dom} = (X_{bg}, X_{dom})$ and $P_{sub} = (X_{bg}, X_{sub})$. We also consider a "corrupted" version of $P_{sub}$, which could be $P_{dom}$ itself or another prompt designed to elicit $Y_{dom}$. Let's denote the "clean" input as $P_{clean}$ (typically $P_{sub}$) and the "corrupted" input as $P_{corr}$ (designed to lead to $Y_{dom}$).

    \item \textbf{Activation Difference Calculation:} We run both $P_{clean}$ and $P_{corr}$ through the model. For each node $v \in V$ that is a potential parent in an edge, we record its output activation. The difference in activations between the clean and corrupted runs for a node $v_p$ (parent) is denoted as $\Delta A(v_p) = A_{clean}(v_p) - A_{corr}(v_p)$.

    \item \textbf{Edge Scoring via Gradient-based Attribution:} To score an edge $e = (v_p, v_c)$ (from parent $v_p$ to child $v_c$), we focus on how patching the activation from $v_p$ (i.e., using $A_{clean}(v_p)$ instead of $A_{corr}(v_p)$ when $P_{corr}$ is the main input) affects a chosen metric $\mathcal{M}$. This metric $\mathcal{M}$ is designed to measure the model's tendency towards generating $Y_{sub}$ versus $Y_{dom}$ when the input is $P_{sub}$. A common choice for $\mathcal{M}$ could be the logit difference between $Y_{sub}$ and $Y_{dom}$ at the final layer, or a metric related to our Relative Overshadowing rate ($RO$).

    The score $S(e)$ for an edge $e$ can be approximated by the product of the activation difference from its parent node and the gradient of the metric $\mathcal{M}$ with respect to the input of its child node, when the child node receives the "clean" activation from the parent while other inputs are "corrupted": $$ S(e) \approx \mathbb{E}_{P_{sub}} \left[ \Delta A(v_p) \cdot \frac{\partial \mathcal{M}(Y_{target} | P_{sub})}{\partial A_{input}(v_c)} \right] $$ where $Y_{target}$ is ideally $Y_{sub}$. The expectation $\mathbb{E}$ is taken over instances of $P_{sub}$ in evaluation set. In practice, methods like Integrated Gradients (IG) are often used to refine this attribution by integrating gradients along a path from a baseline (corrupted) input to the actual (clean) input.

    \item \textbf{Circuit Pruning:} Based on the calculated scores $S(e)$, edges with scores below a certain threshold $\tau$, or alternatively, edges outside the top-N highest scores, are pruned from the graph $G$. The remaining nodes and edges form the knowledge circuit $C_{sub}$. $$ C_{sub} = (V_{sub}, E_{sub}) $$ where $E_{sub} = \{e \in E \mid |S(e)| \geq \tau \}$ (or top-N criterion) and $V_{sub}$ consists of nodes connected by edges in $E_{sub}$.

    This constructed circuit $C_{sub}$ is then analyzed to understand how dominant knowledge $K_{dom}$ might overshadow $K_{sub}$ by examining the attentional features and information flow within it, especially when processing $P_{sub}$.

    \item \textbf{Threshold $\tau$ Optimization:} By constructing a circuit for a given edge pruning threshold, $\tau$, and then evaluating its performance metric, $M$, we can establish the relationship between $\tau$ and $\mathcal{M}$. Therefore, optimizing the circuit is equivalent to finding an optimal threshold, $\tau_{opt}$, that maximizes the performance metric $\mathcal{M}$. In practice, however, we directly optimize the number of edges $n$ rather than the threshold $\tau$. We divide this optimization process into two stages. In the first stage, we start from an initial edge count and incrementally increase the number of edges at uniform intervals to map out the relationship between $\mathcal{M}$ and $n$. We find that this relationship is multi-modal, as shown in Figure \ref{fig:natural}. In the second stage, we employ the golden-section search algorithm within the most promising interval identified previously to precisely locate the optimal number of edges, $n_{opt}$. This process yields the optimized circuit, $C_{opt}(n_{opt})$.

\end{enumerate}

\subsection{Automated Component Identification for Recovery}
\label{app:location}
\textbf{Identifying the Overshadowed Component $X_{sub}^*$.}
A critical precursor to effective circuit-based recovery is the precise identification of the specific component $X_{sub}^*$ within the subordinate prompt $P_{sub}$ that is being overshadowed. This is achieved by adapting the Relative Pointwise Mutual Information (R-PMI) based methodology from \cite{zhang2025law, zhang2024knowledge}.
The process involves:

 Iteratively generating contrastive prompts $P'_{sub}$ by deleting each candidate token $X'_{sub}$ (a potential overshadowed component) from the original $P_{sub}$.
 
  For each pair $(P_{sub}, P'_{sub})$, calculating the R-PMI for tokens $y_i$ in the intersection of their top-$k$ next-token candidate sets, $V_{top}(P_{sub}) \cap V_{top}(P'_{sub})$, using \begin{align*}
R\textnormal{-}PMI(y_i; P_{sub}, P'_{sub}) = {} & \log P(y_i | P_{sub}) \\ 
& - \log P(y_i | P'_{sub}).                                            
\end{align*}
    Summing only the negative R-PMI values to obtain $$S_{R\textnormal{-}PMI}(P_{sub}, P'_{sub}) = \sum \min(R\textnormal{-}PMI(y_i), 0).$$

The $X'_{sub}$ that yields the minimum (most negative) $S_{R\textnormal{-}PMI}$ is identified as the primary overshadowed component, $X_{sub}^*$. This selection is based on the rationale that removing the true $X_{sub}^*$ most strongly exposes the model's bias towards outputs favored by the dominant knowledge pattern. 

In practical implementation, we enhance the identification of $X_{sub}^*$ by calculating a weighted $S_{R\textnormal{-}PMI}$. To do this, we weight each token's $R-PMI(y_i)$ value by the variance of its log probability change, $var(y_i)$, observed when masking different candidates of $X_{sub}$. This results in $ S_{R\textnormal{-}PMI}(P_{sub}, P'_{sub})_{weighted} = \sum \min(R\textnormal{-}PMI(y_i) \cdot Var(y_i), 0)$, which amplifies the effect from the $Y_sub$ token which is most sensitive to the masking of $X^*_{sub}$, enabling a more robust identification.

\textbf{Identifying Target Subordinate Output $Y_{sub}$.}
The intended subordinate output $Y_{sub}$ is identified by assessing which token from $V_{top}(P_{sub})$ (the top-k candidates for the original prompt $P_{sub}=(X_{bg}, X_{sub})$) exhibits the most significant improvement when the overshadowing influence of background knowledge ($X_{bg}$) or other non-subordinate components is mitigated. Specifically, we generate contrastive prompts $P'_{sub}$ by masking or altering components of $X_{bg}$ within $P_{sub}$. $Y_{sub}$ is then the token $y_i \in V_{top}(P_{sub})$ that shows the most substantial rank improvement in these modified prompts $P'_{sub}$ compared to its rank in the original $P_{sub}$. Furthermore, to properly value the rank improvements of initially high-ranked tokens like $Y_{dom}$ and $Y_{sub}$, which have limited room for elevation, we introduce a weight to measure the original rank's impact. This yields the final weighted rank elevation metric for every predicted top-$k$ next-tokens: $ \text{Weighted Rank Elevation} = (k - \text{rank}_{\text{original}}) \times \text{Avg. Elevation of Rank}.$

 This approach preserves the effect of rank improvements for top-ranked tokens. And the rank elevation signifies that the mitigation of impact from $X_{dom} \leftrightarrow X_{bg}$ make overshadowing association weaken, which improves the rank of $Y_{sub}$.

\textbf{Identifying Dominant Output $Y_{dom}$.}
The dominant output $Y_{dom}$ is identified as the token that maintains the highest average rank across all contrastive prompts $P'_{sub}$ generated by deleting different candidate tokens $X'_{sub}$ from $P_{sub}$. This token represents the model's most consistent, default output tendency when specific subordinate cues are variably weakened, likely reflecting the pervasive influence of dominant knowledge associated with the background $X_{bg}$.

With $X_{bg}$ (background knowledge), $X_{sub}$ (identified subordinate component), the expected $Y_{sub}$, and the interfering $Y_{dom}$ established, we prepare the paired inputs required for knowledge circuit construction.
The \textbf{clean input} is the original subordinate prompt $P_{sub} = (X_{bg}, X_{sub})$, for which the desired output is $Y_{sub}$.
To create the \textbf{corrupt input} $P_{dom}$, which is designed to elicit the overshadowing effect and output $Y_{dom}$, we maintain the background knowledge $X_{bg}$ but replace the subordinate component $X_{sub}$ with a generic placeholder token, such as 'something'. Thus, $P_{dom} = (X_{bg}, \text{``something''})$.
This specific formulation of $P_{dom}$ ensures that while the input structure is similar to $P_{sub}$, the absence of $X_{sub}$ allows the strong $X_{bg} \leftrightarrow Y_{dom}$ association to dominate, leading to the incorrect prediction $Y_{dom}$. These paired inputs, $P_{sub}$ and $P_{dom}$, then serve as the foundation for the activation difference calculations in our circuit analysis.

Some more circuit-based overshadowing recovery cases are shown in Table \ref{tab:case}.

\begin{table*}[t]
\caption{Circuit-based overshadowing recovery cases}
\begin{tblr}{
  width = \linewidth,
  colspec = {Q[c,m,wd=0.05\linewidth]   % Group
             Q[l,m,wd=0.23\linewidth]   % Psub with {Xsub}
             Q[l,m,wd=0.12\linewidth]   % M indicator
             Q[c,m,wd=0.1\linewidth]    % Ydom & Ysub
             Q[l,m,wd=0.18\linewidth]   % Full Model Top 5 Prediction
             Q[l,m,wd=0.18\linewidth]}, % Circuit Top 5 Prediction
  % --- 字体设置 ---
  cells = {font=\scriptsize}, % <--- 确保这一行存在并正确
  % --- 特定行/单元格的覆盖 ---
  row{1} = {font+=\bfseries, bg=gray!25}, % 表头在全局 scriptsize 基础上加粗 (font+=)
                                         % 或者使用 font=\bfseries\scriptsize 明确设置
  % --- 背景色 ---
  row{2-6} = {bg=lightgray!50},   % Group 1
  row{7-11} = {bg=white},         % Group 2 (示例)
  row{12-16} = {bg=lightgray!50},
   row{17-21} = {bg=white}, 
  % --- 线条设置 ---
  hline{1,Z} = {0.7pt,solid},
  hline{2} = {0.4pt,solid},
  hline{7} = {0.5pt,solid},       % Group 1 和 Group 2 之间
  hline{12} = {0.5pt,solid},      % Group 2 和 Group 3 之间
   hline{17} = {0.5pt,solid},      % Group 2 和 Group 3 之间
}
    % 表头
    \SetCell[c=1]{c,m} Case
      & \SetCell[c=1]{c,m} $P_{sub}$ with $\{X_{sub}\}$
      & \SetCell[c=1]{c,m} $\mathcal{M}$ indicator (logits difference)
      & \SetCell[c=1]{c,m} $Y_{dom} \; \& \; Y_{sub}$
      & \SetCell[c=1]{c,m} Full Model Top 5 Prediction
      & \SetCell[c=1]{c,m} Circuit Top 5 Prediction \\
    % --- Group 1 ---
    \SetCell[r=5]{c,m} Case 1
      & \SetCell[r=5]{l,m} Analysis of the Chrono-Filter device efficiency for temporal sorting shows outcome \{filtration overload\}
      & \SetCell[r=5]{l,m} Original model:-1.283 \& Circuit:  0.764
      & \SetCell[r=5]{c,m} |Time| \& |Tem|
     & Rank 0: Logit: 16.18 Prob: 32.18\% Token: \texttt{|Tem|}
      & Rank 0: Logit: 21.03 Prob: 73.70\% Token: \texttt{|Time|} \\
       & & & & Rank 1: Logit: 15.42 Prob: 14.98\% Token: \texttt{|Time|}
      & Rank 1: Logit: 19.75 Prob: 20.42\% Token: \texttt{|Tem|}       \\
     & & & &  Rank 2: Logit: 14.21 Prob:  4.47\% Token: \texttt{|Sp|}
      & Rank 2: Logit: 17.34 Prob:  1.84\% Token: \texttt{|Filter|}  \\
      & & & &  Rank 3: Logit: 14.05 Prob:  3.83\% Token: \texttt{|T|}
      &  Rank 3: Logit: 16.40 Prob:  0.72\% Token: \texttt{|E|}      \\
      & & & & Rank 4: Logit: 13.84 Prob:  3.10\% Token: \texttt{|Custom|}
      &  Rank 4: Logit: 16.29 Prob:  0.64\% Token: \texttt{|Sp|}\\
  % --- End of Group 1 ---
    % --- Group 1 ---
    \SetCell[r=5]{c,m} Case 2
      & \SetCell[r=5]{l,m} Constructing psionic wave emitters necessitates precise tuning involving specialized harmonic \{feedback loop\}
      & \SetCell[r=5]{l,m} Original model: -0.745\& Circuit:  2.490
      & \SetCell[r=5]{c,m} |Wave| \& |E|
     &  Rank 0: Logit: 17.64 Prob: 37.18\% Token: \texttt{|Wave|}
      & Rank 0: Logit: 17.96 Prob: 46.34\% Token: \texttt{|E|} \\
       & & & &  Rank 1: Logit: 16.89 Prob: 17.65\% Token: \texttt{|E|}
      & Rank 1: Logit: 16.47 Prob: 10.39\% Token: \texttt{|Ps|}    \\
     & & & &
    Rank 2: Logit: 15.35 Prob:  3.76\% Token: \texttt{|Ps|} &
    Rank 2: Logit: 15.47 Prob:  3.84\% Token: \texttt{|Wave|} \\

    & & & &
    Rank 3: Logit: 15.29 Prob:  3.57\% Token: \texttt{|St|} &
    Rank 3: Logit: 15.31 Prob:  3.28\% Token: \texttt{|Energy|} \\

    & & & &
    Rank 4: Logit: 15.15 Prob:  3.09\% Token: \texttt{|emitter|} &
    Rank 4: Logit: 14.89 Prob:  2.15\% Token: \texttt{|emitter|} \\

     % --- Group 1 ---
    \SetCell[r=5]{c,m} Case 3
      & \SetCell[r=5]{l,m} Analyzing Ectoplasmic Conduit energy transfer efficiency through degrading \{structure reinforcement reveals\}
      & \SetCell[r=5]{l,m} Original model: -3.019 \& Circuit: 0.523
      & \SetCell[r=5]{c,m} |Transfer| \& |Emit|
     & Rank 0: Logit: 49.08 Prob: 49.26\% Token: \texttt{|Transfer|}
      & Rank 0: Logit: 44.54 Prob: 47.18\% Token: \texttt{|Emit|} \\
        & & & &
    Rank 1: Logit: 48.68 Prob: 33.09\% Token: \texttt{|St|} &
    Rank 1: Logit: 44.01 Prob: 27.96\% Token: \texttt{|Transfer|} \\

    & & & &
    Rank 2: Logit: 46.66 Prob:  4.38\% Token: \texttt{|Ada|} &
    Rank 2: Logit: 42.38 Prob:  5.48\% Token: \texttt{|St|} \\

    & & & &
    Rank 3: Logit: 46.37 Prob:  3.28\% Token: \texttt{|Flow|} &
    Rank 3: Logit: 42.24 Prob:  4.75\% Token: \texttt{|Energy|} \\

    & & & &
    Rank 4: Logit: 46.06 Prob:  2.41\% Token: \texttt{|Emit|} &
    Rank 4: Logit: 41.64 Prob:  2.61\% Token: \texttt{|Mi|} \\

       % --- Group 4 ---
    \SetCell[r=5]{c,m} Case 4
      & \SetCell[r=5]{l,m} Shard Relic residual energy output response to sudden energy \{conduit field spikes\} shows
      & \SetCell[r=5]{l,m} Original model: -2.623 \& Circuit: 3.202
      & \SetCell[r=5]{c,m} |Output| \& |St|
     &  Rank 0: Logit: 40.96 Prob: 91.94\% Token: \texttt{|Output|}
      & Rank 0: Logit: 34.61 Prob: 51.02\% Token: \texttt{|St|} \\
     & & & &
    Rank 1: Logit: 38.34 Prob:  6.67\% Token: \texttt{|St|} &
    Rank 1: Logit: 34.24 Prob: 35.19\% Token: \texttt{|Field|} \\

    & & & &
    Rank 2: Logit: 35.87 Prob:  0.56\% Token: \texttt{|Fl|} &
    Rank 2: Logit: 31.41 Prob:  2.08\% Token: \texttt{|Output|} \\

    & & & &
    Rank 3: Logit: 35.82 Prob:  0.54\% Token: \texttt{|Trans|} &
    Rank 3: Logit: 30.97 Prob:  1.34\% Token: \texttt{|Har|} \\

    & & & &
    Rank 4: Logit: 33.51 Prob:  0.05\% Token: \texttt{|Un|} &
    Rank 4: Logit: 30.84 Prob:  1.18\% Token: \texttt{|Fl|} \\
    % --- End of Case 4 ---
\end{tblr}
\label{tab:case} % 保持你的标签
\vspace{-0.1cm}
\end{table*}

\section{Dataset Details}
\label{app:dataset}

\begin{figure*}[t]
  \centering
  \begin{subfigure}[b]{0.45\linewidth}
    \includegraphics[width=\linewidth]{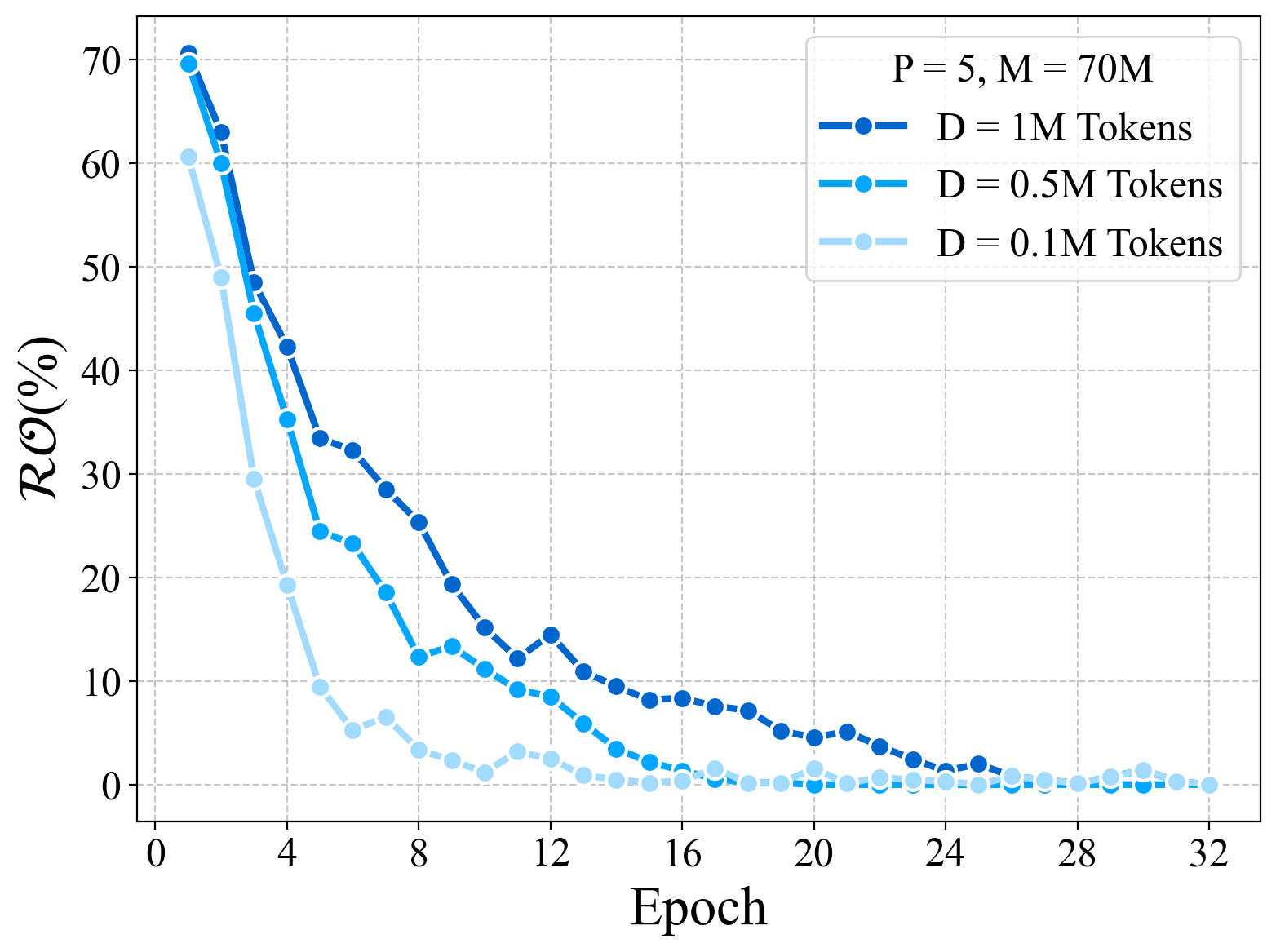} % 替换为你的图片路径
    \caption{The $\mathcal{RO}$ during training phase of  the different fine-tuning dataset size (D).}
    \label{fig:ro_ft_d}
  \end{subfigure}
  \hfill % 子图间距
  \begin{subfigure}[b]{0.45\linewidth}
    \includegraphics[width=\linewidth]{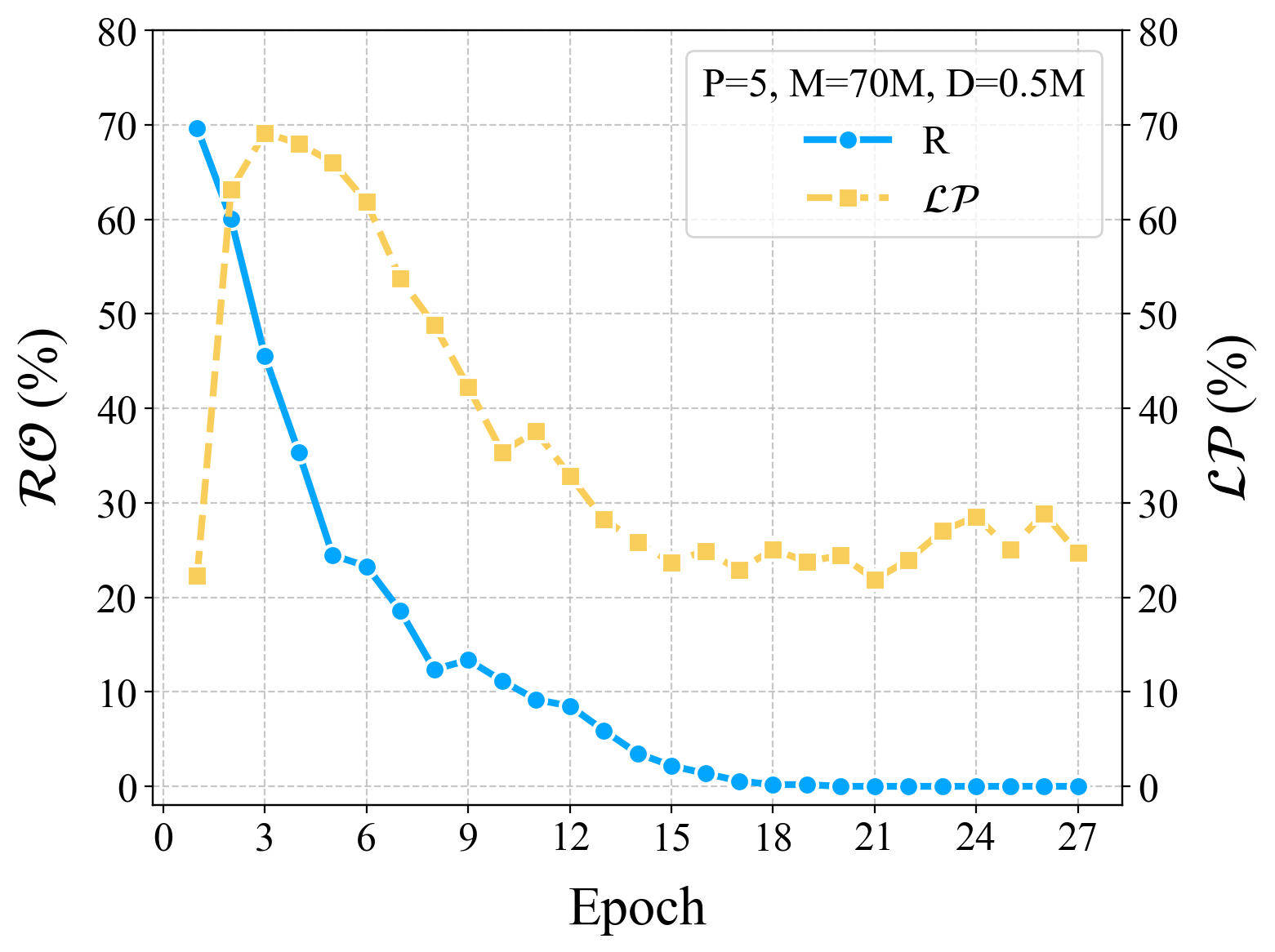} % 替换为你的图片路径
    \caption{The co-evolution of $\mathcal{LP}$ and $\mathcal{RO}$ during training phase on fine-tuning dataset.}
    \label{fig:ro_dt_lp}
    
  \end{subfigure}
\vspace{-0.2cm}
  \caption{The dynamics analysis of knowledge overshadowing in fine-tuning dataset.}

  \label{fig:ftdata}
    \vspace{-0.5cm}
\end{figure*}

\subsection{Detailed Synthetic Dataset Construction}

The synthetic dataset was constructed through the following steps to ensure controlled conditions for analyzing knowledge overshadowing dynamics:

   \textbf{Fixing text lengths.} For all generated data instances, consistent token lengths are maintained. The background knowledge ($X_{bg}$) was set to a length of 4 tokens. All other core components, namely the dominant knowledge entity ($X_{dom}$), subordinate knowledge entity ($X_{sub}$), dominant output ($Y_{dom}$), and subordinate output ($Y_{sub}$), are each set to a length of 1 token.

    \textbf{Dataset generation for specific D and P Combinations.} For each defined combination of total dataset size (D) and knowledge popularity (P), the dataset was built as follows:
    The dataset comprises multiple distinct groups of knowledge instances. Each group consists of P+1 knowledge prompts: a set of P dominant knowledge prompts $\{P_{dom}^1, P_{dom}^2, \dots, P_{dom}^P\}$ and one subordinate knowledge prompt $P_{sub}$.
    Within each group:
    \begin{itemize}
        \item For the P dominant prompts, the actual dominant knowledge entities $\{X_{dom}^1, X_{dom}^2, \dots, X_{dom}^P\}$ are all unique. However, they all share the \textit{same} background knowledge component ($X_{bg}$) and are associated with the \textit{same} dominant output ($Y_{dom\_g}$). Thus, each $P_{dom}^i = (X_{bg}, X_{dom}^i)$ is paired with $Y_{dom}^g$.
        \item The single subordinate prompt $P_{sub} = (X_{bg}, X_{sub})$ uses the \textit{same} background knowledge $X_{bg}$ as the dominant prompts in that group. However, its subordinate knowledge entity $X_{sub}$ is distinct from all $X_{dom}^i$ entities in that group, and its corresponding output $Y_{sub}$ is distinct from any $Y_{dom}$ in group. 
    \end{itemize}
    This\;structure\;creates\;a\;group: $\{(P_{dom}^1, Y_{dom}^1), \dots, ((P_{dom}^P, Y_{dom}^P), (P_{sub}, Y_{sub})\}$.

    Multiple such groups are generated. All tokens for $X_{bg}, X_{dom}^i, X_{sub}, Y_{dom}, Y_{sub}$ within each group, and across different groups, are randomly sampled from the Pythia tokenizer vocabulary, ensuring no overlap between the core entities of different groups. This process was repeated until the total number of tokens in the dataset reached the target size D.

\textbf{Cases illustration.} We illustrate some groups for P=5 dataset in Table \ref{tab:syndata}. We directly show token id.

\begin{table*}[t!]
  \centering
  \caption{Illustrative examples from the synthetic dataset (P=5). Each data entry is a row, with fine lines separating entries within a group. Token IDs are shown. }
  \label{tab:synthetic_data_detailed_scaled}
  \scriptsize % <--- 将表格内所有字体设置为 \scriptsize (约 70%)
             % 你也可以尝试 \footnotesize (约 80%)
  \begin{tblr}{
    width = \linewidth, % 尝试让表格宽度等于行宽
    colspec = {Q[c,m,wd=0.1\linewidth]   % Group
               Q[l,m,wd=0.26\linewidth]   % Xbg (给Xbg更多空间以容纳token列表)
               Q[c,m,wd=0.12\linewidth]  % Xdom
               Q[c,m,wd=0.12\linewidth]  % Ydom
               Q[c,m,wd=0.12\linewidth]  % Xsub
               Q[c,m,wd=0.12\linewidth]}, % Ysub
               % 总宽度: 0.1 + 0.3 + 0.12*4 = 0.1 + 0.3 + 0.48 = 0.88\linewidth。留有列间距。
    row{1} = {font=\bfseries\scriptsize, bg=gray!25}, % 表头也用 scriptsize 粗体
    % --- 为 Group 1 和 Group 3 的行设置背景色 ---
    row{2-7} = {bg=lightgray},   % Group 1
    row{14-19} = {bg=lightgray}, % Group 3
    % --- 线条设置 ---
    hline{1,Z} = {0.7pt,solid}, % 表格顶线和底线 (线宽略微调整以适应字体)
    hline{2} = {0.4pt,solid},   % 表头下的线
    % ---- 组与组之间用稍粗的线 ----
    hline{8} = {0.5pt,solid},   % Group 1 和 Group 2 之间
    hline{14} = {0.5pt,solid},  % Group 2 和 Group 3 之间
    % ---- 组内部逻辑行之间用非常细的实线 ----
    % hline{3-7} = {0.1pt, , gray},
    % hline{9-13} = {0.1pt, , gray},
    % hline{15-19} = {0.1pt, , gray},
  }
    % 表头
    \SetCell[c=1]{c,m} Group
      & \SetCell[c=1]{c,m} $X_{bg}$
      & \SetCell[c=1]{c,m} $X_{dom}$
      & \SetCell[c=1]{c,m} $Y_{dom}$
      & \SetCell[c=1]{c,m} $X_{sub}$
      & \SetCell[c=1]{c,m} $Y_{sub}$ \\
    % --- Group 1 ---
    \SetCell[r=6]{c,m} Group 1
      & $[10030, 16936, 1050, 10565]$ & $10279$ & $20730$ &         &        \\
      & $[10030, 16936, 1050, 10565]$ & $24327$ & $20730$ &         &        \\
      & $[10030, 16936, 1050, 10565]$ & $4619$  & $20730$ &         &        \\
      & $[10030, 16936, 1050, 10565]$ & $5137$  & $20730$ &         &        \\
      & $[10030, 16936, 1050, 10565]$ & $785$   & $20730$ &         &        \\
      & $[10030, 16936, 1050, 10565]$ &         &         & $18941$ & $3519$ \\
    % --- Group 2 ---
    \SetCell[r=6]{c,m} Group 2
      & $[17026, 8837, 3802, 28741]$ & $2496$  & $1077$  &         &        \\
      & $[17026, 8837, 3802, 28741]$ & $3530$  & $1077$  &         &        \\
      & $[17026, 8837, 3802, 28741]$ & $11948$ & $1077$  &         &        \\
      & $[17026, 8837, 3802, 28741]$ & $2028$  & $1077$  &         &        \\
      & $[17026, 8837, 3802, 28741]$ & $9389$  & $1077$  &         &        \\
      & $[17026, 8837, 3802, 28741]$ &         &         & $25814$ & $5374$ \\
    % --- Group 3 ---
    \SetCell[r=6]{c,m} Group 3
      & $[18131, 14501, 21161, 311]$ & $4706$  & $7790$  &         &        \\
      & $[18131, 14501, 21161, 311]$ & $778$   & $7790$  &         &        \\
      & $[18131, 14501, 21161, 311]$ & $18762$ & $7790$  &         &        \\
      & $[18131, 14501, 21161, 311]$ & $28591$ & $7790$  &         &        \\
      & $[18131, 14501, 21161, 311]$ & $28981$ & $7790$  &         &        \\
      & $[18131, 14501, 21161, 311]$ &         &         & $8447$  & $6129$ \\
  \end{tblr}
  \label{tab:syndata}
\end{table*}

\subsection{fine-tuning dataset} 

For the fine-tuning dataset, we utilized the Qwen-Long API to generate instances of virtual knowledge.
This generated data subsequently underwent manual review to identify and remove any instances that are overly repetitive or semantically too similar, ensuring a degree of diversity, resulted in D = 1M.

A key distinction from the synthetic dataset construction is that we did not strictly control token lengths for each component in this dataset.
Instead of randomly sampled token IDs, the fine-tuning dataset consists of actual linguistic statements that, while syntactically and semantically coherent, represent virtual (i.e., fabricated but plausible) knowledge.
The underlying pattern of dominant and subordinate knowledge construction, however, mirrors that of the synthetic dataset.

As an example of this dataset, we set the knowledge popularity P=5. Some illustrative cases from the dataset are shown in Table \ref{tab:ftdata}.

\begin{table*}[t!]
  \centering
  \caption{Illustrative examples from the fine-tuning Dataset (P=5). Each data entry is a row, with fine lines separating entries within a group. }
  \label{tab:ftdata_detailed_scaled_3groups}
  \scriptsize % 表格内所有字体设置为 \tiny
  \begin{tblr}{
    width = \linewidth,
    % colspec = {Q[c,m,wd=0.1\linewidth] % Group 列
    %            X[l,m]                  % Xbg (Xshare)
    %            X[0.8,l,m]              % Xdom (Xa) / Xsub (Xb)
    %            X[0.7,l,m]              % Ydom (Ya) / Ysub (Yb)
    %            % 上面三列的权重可以调整，确保内容能放下
    %            % 移除了原先为Xsub, Ysub单独列的colspec，因为现在数据都在 Xdom/Xsub 和 Ydom/Ysub 列中
    %            % 如果你需要单独列出Xsub和Ysub，需要重新设计列
    %            % 这里假设Xa/Xb放在同一列，Ya/Yb放在同一列
    %            % 因此表格现在是3列数据 + 1列Group
    %            % 调整为：Group | Xshare (Xbg) | Xa/Xb (Xdom/Xsub) | Ya/Yb (Ydom/Ysub)
    %            % 如果严格按照之前的Xbg, Xdom, Ydom, Xsub, Ysub五列，需要将数据拆分
    %           },
    % 重新定义colspec以匹配你的新数据结构 (Xshare, Xa/Xb, Ya/Yb)
    % 并且保持与之前表格类似的列数和用途
    colspec = {Q[c,m,wd=0.08\linewidth]   % Group
               X[1.5,l,m]                % Xbg (Xshare)
               X[l,m]                    % Xdom (Xa)
               X[0.8,l,m]                % Ydom (Ya)
               X[l,m]                    % Xsub (Xb)
               X[0.8,l,m]},              % Ysub (Yb)
    row{1} = {font=\bfseries, bg=gray!25}, % 表头
    % --- 为 Group 1 和 Group 3 的行设置背景色 ---
    row{2-7} = {bg=lightgray},    % Group 1 (Starfire)
    % Group 2 (Chronos) 无特殊背景色 (行 8-13)
    row{14-19} = {bg=lightgray},  % Group 3 (Xylosian)
    % --- 线条设置 ---
    hline{1,Z} = {0.6pt,solid},
    hline{2} = {0.3pt,solid},
    hline{8,14} = {0.4pt,solid}, % 组间分隔线
    % % ---- 组内部逻辑行之间用非常细的点线 ----
    % hline{3-7} = {0.1pt, dotted, gray},   % Group 1 内
    % hline{9-13} = {0.1pt, dotted, gray},  % Group 2 内
    % hline{15-19} = {0.1pt, dotted, gray}, % Group 3 内
  }
    % 表头 - 保持之前的5列数据结构
    \SetCell[c=1]{c,m} Group
      & \SetCell[c=1]{c,m} $X_{bg}$ 
      & \SetCell[c=1]{c,m} $X_{dom}$ 
      & \SetCell[c=1]{c,m} $Y_{dom}$ 
      & \SetCell[c=1]{c,m} $X_{sub}$
      & \SetCell[c=1]{c,m} $Y_{sub}$  \\
    % --- 第一个逻辑大组 (Starfire Crystal Engine) ---
    \SetCell[r=6]{c,m} Group 1 % (Starfire Grp)
      & Analysis of the Starfire Crystal Engine reveals primary energy & output peak resonance & Pure Nova &  &  \\
      & Analysis of the Starfire Crystal Engine reveals primary energy & output idle cycle & Pure Nova &  &  \\
      & Analysis of the Starfire Crystal Engine reveals primary energy & output phase synchronicity & Pure Nova &  &  \\
      & Analysis of the Starfire Crystal Engine reveals primary energy & output null gravity & Pure Nova &  &  \\
      & Analysis of the Starfire Crystal Engine reveals primary energy & output core temperature & Pure Nova &  &  \\
      & Analysis of the Starfire Crystal Engine reveals primary energy &  &  & output cold ignition & Quantum Flux \\
    % --- 第二个逻辑大组 (Chronos Orb) ---
    \SetCell[r=6]{c,m} Group 2 % (Chronos Grp)
      & Observation of the Chronos Orb's temporal distortion signature & indicates full moonlight & Stable Chrono-displacement &  &  \\
      & Observation of the Chronos Orb's temporal distortion signature & indicates near Leyline & Stable Chrono-displacement &  &  \\
      & Observation of the Chronos Orb's temporal distortion signature & indicates while attuned & Stable Chrono-displacement &  &  \\
      & Observation of the Chronos Orb's temporal distortion signature & indicates during solstice & Stable Chrono-displacement &  &  \\
      & Observation of the Chronos Orb's temporal distortion signature & indicates etherium interaction & Stable Chrono-displacement &  &  \\
      & Observation of the Chronos Orb's temporal distortion signature &  &  & indicates sudden impact & Temporal Anomaly \\
    % --- 第三个逻辑大组 (Xylosian biology) ---
    \SetCell[r=6]{c,m} Group 3 % (Xylosian Grp)
      & Study of the Xylosian biology reveals primary nutrient absorption & method consuming light & Molecular Assimilation & & \\
      & Study of the Xylosian biology reveals primary nutrient absorption & method after sonic bath & Molecular Assimilation & & \\
      & Study of the Xylosian biology reveals primary nutrient absorption & method during digestion & Molecular Assimilation & & \\
      & Study of the Xylosian biology reveals primary nutrient absorption & method high pressure & Molecular Assimilation & & \\
      & Study of the Xylosian biology reveals primary nutrient absorption & method thermal vent & Molecular Assimilation & & \\
      & Study of the Xylosian biology reveals primary nutrient absorption & & & method xenoflora consumption & Crystalline Excretion \\
  \end{tblr}
  \label{tab:ftdata}
\end{table*}

%%%%%%%%%%%%%%%%%%%%%%%%%%%%%%
\begin{figure*}[t]
  \centering
  \begin{subfigure}[b]{0.45\linewidth}
    \includegraphics[width=\linewidth]{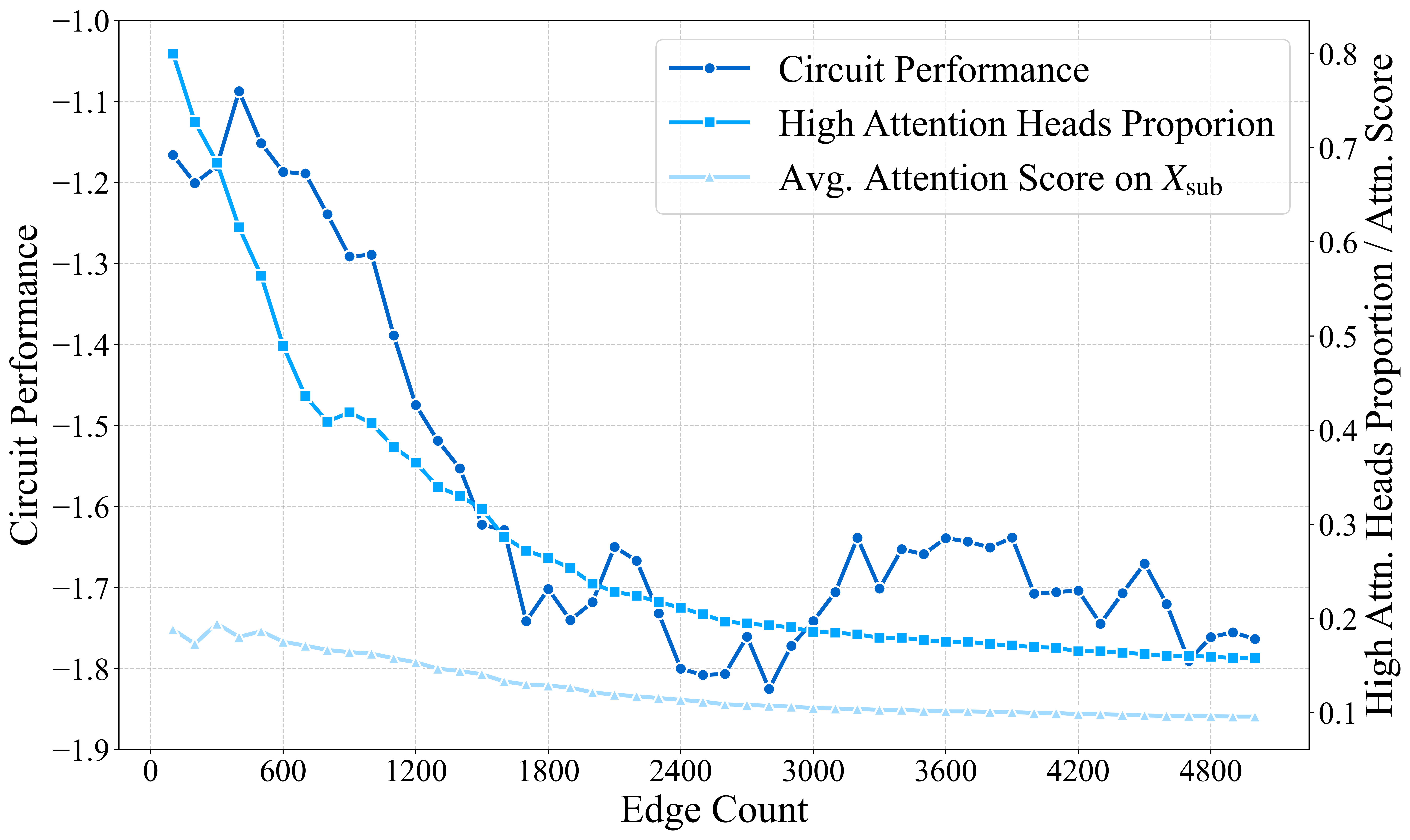} % 替换为你的图片路径
    \caption{gpt2-medium}
    \label{fig:gpt_natural}
  \end{subfigure}
  \hfill % 子图间距
  \begin{subfigure}[b]{0.45\linewidth}
    \includegraphics[width=\linewidth]{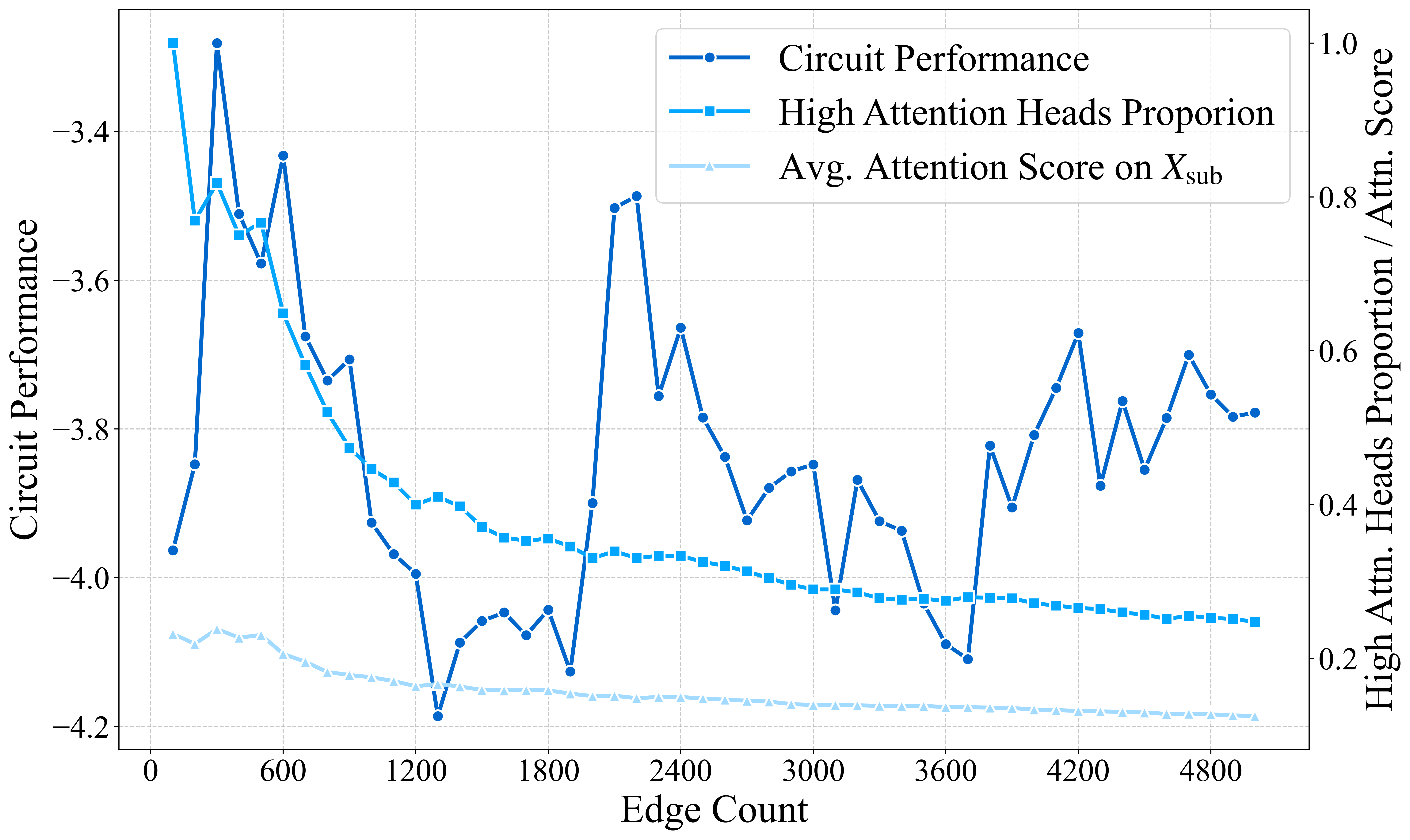} % 替换为你的图片路径
    \caption{pythia-410m}
    \label{fig:pythia_natural}
    
  \end{subfigure}
\vspace{-0.2cm}
  \caption{The relationship between circuit performance, high-attention heads proportion, average attention score on $X_{sub}$ and the number of edges in gpt2-medium and pythia-410m.
}

  \label{fig:natural}
    \vspace{-0.5cm}
\end{figure*}

%%%%%%%%%%%%%%%%%%%%%%%%%%%%%%%%

\section{More Dynamics Analysis on fine-tuning Dataset}
\label{app:dynamics}

In addition to validating the efficacy of our circuit-based overshadowing recovery method, the fine-tuning dataset serves a dual purpose. We also leverage it to empirically verify our conclusions regarding the training dynamics of knowledge overshadowing, specifically concerning the impact of Dataset Size (D).
Consistent with our dynamic analysis findings, we investigate whether a larger D indeed correlates with a slower recovery rate from knowledge overshadowing and a prolonged duration of the hallucination effect.
To this end, we conduct experiments on the fine-tuning dataset by fixing Knowledge Popularity at P=5 and Model Size at M=70M, while varying D across values of 0.1M, 0.5M, and 1M tokens. The results, as depicted in Figure \ref{fig:ro_ft_d}, corroborate this relationship.
Furthermore, under the specific configuration of P=5, M=70M, and D=0.5M on the fine-tuning dataset, we re-examine the interplay between the loss proportion of subordinate knowledge ($\mathcal{LP}$) and the relative overshadowing rate ($\mathcal{RO}$). As shown in Figure \ref{fig:ro_dt_lp}, the observations again support the hypothesis that insufficient optimization of subordinate knowledge contributes to the persistence of knowledge overshadowing.

It is noteworthy that distinct behaviors are observed when comparing the fine-tuning dataset to the synthetic dataset.
Firstly, the recovery from overshadowing on the fine-tuning dataset is generally slower than on the synthetic dataset for same D. This can be attributed to the richer semantic relationships and greater complexity inherent in the natural language of the fine-tuning data, which presents a more challenging learning task.

Secondly, we observe that the fine-tuning dataset exhibits a minimal or absent onset phase for knowledge overshadowing, where $\mathcal{RO}$ typically rise. This is because fine-tuning commences from a pre-trained model, which has already moved beyond the initial epochs of chaotic, random predictions. Consequently, the model can very rapidly generalize strong association patterns present in the fine-tuning data.
Moreover, the diverse and varied forms of data within the fine-tuning set may act akin to a beneficial noise signal, prompting the model to pay closer attention to distinguishing features and differences. This inherent data diversity can help preemptively mitigate or even eliminate the early onset stage of knowledge overshadowing that might otherwise be observed.

\section{More Analysis on Natural Language Data}
\label{app:natural}
To further validate the generalization of our findings, we conduct experiments on the pre-trained Pythia-410m and gpt2-medium models using natural language prompts. The paired inputs for this test are: \{$P_{dom}$: "The name of this North Korean politician is", $Y_{dom}$: "Kim Jong Un", $X_{dom}$: "politician"\} and \{$P_{sub}$: "The name of this North Korean singer is", $Y_{sub}$: "Hyon Song-wol", $X_{sub}$: "singer"\}. Since it is hard to reproduce the pre-training process of these models, we instead used the number of edges in the circuit as the independent variable. We investigated the relationship among three metrics: circuit performance, the proportion of high-attention heads, and the average attention score on $X_{sub}$.

The results are shown in Figure \ref{fig:natural}, which lead to the following conclusions:
\begin{itemize}
    \item The model's overall attention to the different parts $\{X_{dom}, X_{sub}\}$ increases as the number of edges is reduced through optimization. Correspondingly, the model's performance improves with this rise in attention. This validates our conclusion that higher attention on $\{X_{dom}, X_{sub}\}$ is key to mitigating knowledge overshadowing.
    
    \item The proportion of high-attention heads also increases during edge optimization. This supports the conclusion that retaining these high-attention heads plays a crucial role in directing the model's focus toward $\{X_{dom}, X_{sub}\}$.
    
    \item The model's performance exhibits a notable multi-modal relationship with the number of edges, which makes the optimization of the circuit's edge count a more complex task.
\end{itemize}

Additionally, our preliminary exploration suggests that knowledge overshadowing can also occur in multi-hop reasoning. When the pre-trained OLMo-7B-Instruct model is given the prompt: "The old country of this famous scientist who is not famous for theory of relativity is", it outputs "Germany". Furthermore, with the prompt: "The name of this scientist who is not famous for theory of relativity is", the model outputs "Albert". We can infer from this that the model still defaults to "Albert Einstein" as the basis for its multi-hop reasoning.

\end{document}